\documentclass[10pt,twocolumn,letterpaper]{article}

\usepackage{cvpr}              

\definecolor{cvprblue}{rgb}{0.21,0.49,0.74}
\usepackage[pagebackref,breaklinks,colorlinks,allcolors=cvprblue]{hyperref}
\usepackage{amsmath} 
\usepackage{amssymb}
\usepackage{multirow}
\usepackage{algorithm}
\usepackage{algpseudocode}
\usepackage{xcolor}
\usepackage{times}
\usepackage[accsupp]{axessibility}
\definecolor{myorange}{RGB}{231,118,41}
\definecolor{mypurple}{RGB}{112,48,160}

\def\paperID{980} 
\def\confName{CVPR}
\def\confYear{2025}

\title{VTON 360: High-Fidelity Virtual Try-On from Any Viewing Direction}

\author{
Zijian He\textsuperscript{1}
\
Yuwei Ning\textsuperscript{1}
\
Yipeng Qin\textsuperscript{2}
\
Guangrun Wang\textsuperscript{1}
\
Sibei Yang\textsuperscript{3}
\
Liang Lin\textsuperscript{1,4,5}
\
Guanbin Li\textsuperscript{1,4,5$^*$}
\\
\textsuperscript{1}Sun Yat-sen University \qquad  \textsuperscript{2}Cardiff University  \qquad \textsuperscript{3}ShanghaiTech University \\  \textsuperscript{4}Guangdong Key Laboratory of Big Data Analysis and Processing \qquad \textsuperscript{5}Peng Cheng Laboratory \\
{\tt\small hezj39@mail2.sysu.edu.cn, yuwei\_ning@hust.edu.cn,\{qinyipeng1991,wanggrun\}@gmail.com,}\\
\vspace{-4pt}
{\tt\small yangsb@shanghaitech.edu.cn linliang@ieee.org, liguanbin@mail.sysu.edu.cn}
}

\begin{document}
\twocolumn[{
\maketitle
\vspace{-10pt}
\begin{center}
    \captionsetup{type=figure}
    \vspace{-22pt}
    \includegraphics[width=1.0\textwidth]{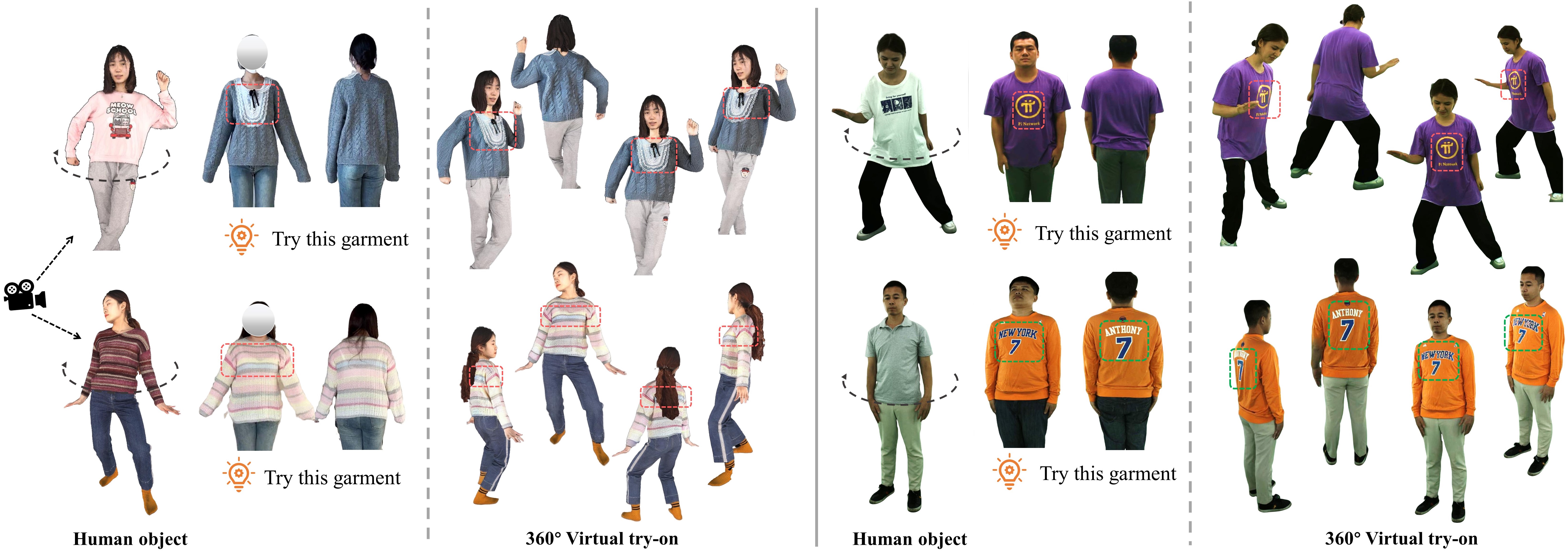}
    \vspace{-15pt}
    \captionof{figure}{\small{\textbf{Results of VTON 360.} Our VTON 360 enables high-fidelity 3D Virtual Try-On (VTON) by seamlessly adapting E-commerce garments onto a clothed 3D human model, supporting full 360$^\circ$ view rendering. 
    The highlighted bounding boxes (dashed line) demonstrate our method's ability to preserve intricate clothing details and patterns (\eg, collar accessories, horizontal line patterns, logos, texts, numbers) across diverse garment types.
    }}
    \vspace{-5pt}
\end{center}
}]
\maketitle
\def\thefootnote{*}\footnotetext{Corresponding author is Guanbin Li.}
\begin{abstract}
Virtual Try-On (VTON) is a transformative technology in e-commerce and fashion design, enabling realistic digital visualization of clothing on individuals.
In this work, we propose VTON 360, a novel 3D VTON method that addresses the open challenge of achieving high-fidelity VTON that supports any-view rendering.
Specifically, we leverage the {\it equivalence} between a 3D model and its rendered multi-view 2D images, and reformulate 3D VTON as an extension of 2D VTON that ensures 3D consistent results across multiple views.
To achieve this, we extend 2D VTON models to include multi-view garments and clothing-agnostic human body images as input, and propose several novel techniques to enhance them, including: i) a pseudo-3D pose representation using normal maps derived from the SMPL-X 3D human model, ii) a multi-view spatial attention mechanism that models the correlations between features from different viewing angles, and iii) a multi-view CLIP embedding that enhances the garment CLIP features used in 2D VTON with camera information. Extensive experiments on large-scale real datasets and clothing images from e-commerce platforms demonstrate the effectiveness of our approach. Project page: \url{https://scnuhealthy.github.io/VTON360}.
\vspace{-10pt}

\end{abstract}

\section{Introduction}
\label{sec:intro}



Virtual Try-On (VTON) enables realistic digital visualization of clothing on individuals and has emerged as a transformative technology in e-commerce and fashion design.
While significant research efforts have been made on 2D VTON solutions~\cite{morelli2023ladi, gou2023taming, rombach2022high, choi2021viton, he2024wildvidfit}, these approaches are inherently limited in their representation of view-related features.
To overcome this limitation and enable high-fidelity any-view rendering, 3D VTON methods were introduced. 

3D VTON requires accurate garment transfer onto a 3D human body while ensuring realistic garment fitting, texture preservation, and 3D consistency. The two primary aims of 3D VTON are i) achieving {\it high-fidelity} and ii) supporting {\it any-view rendering}.
Leveraging the inherent capability of 3D models for {\it any-view rendering}, early 3D VTON methods~\cite{hahn2014subspace,guan2012drape,lahner2018deepwrinkles} make clothing simulation on synthetic human bodies.
Specifically, these methods utilized 3D scanners to capture clothing meshes, followed by the development of specialized dressing algorithms. 
Although effective, these methods rely on costly 3D scanning equipment and the physical presence of the human body/clothing (\ie, not fully virtual), restricting their practicality in real-world applications.
As a byproduct, most early methods focused on developing geometrically correct dressing algorithms using standard templates of human body and clothing models.
Addressing this limitation, researchers extended 3D VTON by introducing algorithms that reconstruct 3D clothing models from input images, enabling the use of image-based clothing inputs~\cite{bhatnagar2019multi,mir2020learning,santesteban2021self,santesteban2022ulnef}.
However, since input clothing images (usually frontal) are inherently 2D and lack multi-view information, this approach struggles to reconstruct high-fidelity clothing models that can be rendered well from all viewing directions.

To complement this missing information, DreamVTON~\cite{xie2024dreamvton} introduces a novel approach that leverages Text-to-Image (T2I) diffusion models to reconstruct both the human body and clothing from input images.
Its key insight is that T2I models learned view-agnostic ``concepts'' of both bodies and garments during their training, and that the corresponding concepts for the input body and clothing images can be obtained using LoRA~\cite{hu2021lora}.
By utilizing Score Distillation Sampling (SDS)~\cite{poole2022dreamfusion}, DreamVTON can generate visual-pleasing 3D VTON results by ensuring consistency between renderings from arbitrary viewpoints and the concepts.
Nonetheless, DreamVTON's high flexibility comes at the cost of low fidelity.
This limitation stems from the fact that the concepts learned by T2I models are semantic in nature, thus lacking 3D geometric consistency and pixel-level accuracy with respect to the input body and clothing images.
Recently, a concurrent work, namely GaussianVTON~\cite{chen2024gaussianvton}, partially addressed this limitation by formulating 3D VTON as a 3D scene editing task, where a given 3D human model is edited using multi-view images generated by 2D VTON methods.
While it significantly enhances the fidelity of the human body, the fidelity and 3D consistency of clothing remain problematic, as there are no 2D VTON methods that can generate multi-view images with 3D consistency. 
Therefore, to the best of our knowledge, achieving high-fidelity 3D VTON that supports any-view rendering remains an open challenge.

In this work, we address the above-mentioned challenge via proposing VTON 360, a novel 3D VTON method that achieves high-fidelity VTON from arbitrary viewing directions.
Similar to GaussianVTON
~\cite{chen2024gaussianvton}, our method edits a given 3D human model by inpainting the rendered images using a latent diffusion model. However, we set ourselves apart through our novel garment fidelity preservation strategy that can generate high-fidelity on-body garments in all viewing directions.
Specifically, we first extend both the garment and clothing-agnostic human body inputs to typical 2D VTON models to leverage multi-view information, including paired front and back view garment images as well as a set of multi-view clothing-agnostic human body images sampled from random azimuth angles.
Then, we propose several novel enhancements to bridge the gap between typical 2D VTON methods and our multi-view 3D consistency requirements: 
i) We propose a pseudo-3D pose representation using normal maps derived from the SMPL-X 3D human model, which captures fine-grained surface orientation details and provides more consistent geometry across views compared to the 2D pose representations (semantic segmentation maps) used in 2D VTON models. 
ii) We design a Multi-view Spatial Attention mechanism that models the correlations between features from different viewing angles, featuring a novel ``correlation'' matrix modeling the relationships among different input views.
iii) We propose a multi-view CLIP embedding that enhances the garment CLIP embedding used in 2D VTON methods with camera information, thereby facilitating network learning of features relevant to a particular view.
Together, these innovations enable our 2D VTON model to generate high-quality, multi-view and 3D-consistent virtual try-on results.
Extensive experiments on Thuman2.0~\cite{yu2021function4d} and MVHumanNet~\cite{xiong2024mvhumannet} datasets demonstrate that our method achieves high fidelity 3D VTON which supports any-view rendering. 
In addition, we show the effectiveness and generalizability of our methodology by testing it using garments from e-commerce platforms.
Our conclusions include:
\begin{itemize}
    \item We propose a novel 3D Virtual Try-On (VTON) method, namely {\it VTON 360}, which achieves high-fidelity VTON from arbitrary viewing directions.
    \item Leveraging the {\it equivalence} between a 3D model and its rendered multi-view 2D images, we reformulate 3D VTON as an extension of 2D VTON that ensures consistent results across multiple views. Specifically, we introduce several novel techniques, including: (i) pseudo-3D pose representation; (ii) multi-view spatial attention; and (iii) multi-view CLIP embedding. These innovations enhance traditional 2D VTON models to generate multi-view and 3D-consistent results.
    \item Extensive experimental results on two large real datasets as well as real clothing images from e-commerce platforms demonstrate the effectiveness of our approach.
\end{itemize}

\section{Related Work}

\noindent\textbf{2D Virtual Try-On. }
2D Virtual Try-On (VTON) aims to transfer a desired garment to the corresponding region of a target human image while preserving the human pose and identity. 
Early methods~\cite{han2018viton,choi2021viton,ge2021parser,dong2022dressing,He_2022_CVPR,Yang_2022_CVPR,lee2022hrviton,bai2022single,men2020controllable,zhang2021pise,ren2022neural} use Generative Adversarial Networks (GANs) to deform the garments to match the target body shape, which a critical step for achieving realistic VTON. However, accurately adapting to diverse real-world conditions remains a significant challenge.
Addressing this issue, recent methods~\cite{morelli2023ladi,gou2023taming,zhu2023tryondiffusion, he2024wildvidfit} reframe 2D VTON as a conditioned inpainting task, leveraging the strong priors provided by diffusion models~\cite{sohl2015deep,song2019generative,ho2020denoising} to achieve promising results.
This strategy is further improved by~\cite{kim2023stableviton, xu2024ootdiffusion, choi2024improving}, which introduce a ReferenceNet to extract hierarchical garment features and apply attention mechanisms to condition the Main UNet.

\vspace{2mm}
\noindent\textbf{3D Virutal Try-On.} 
For 3D Virtual Try-On (VTON), traditional methods~\cite{bridson2002robust,guan2012drape,hahn2014subspace,lahner2018deepwrinkles,pons2017clothcap} rely on 3D scanning or cloth simulation to generate highly precise body and garment geometry. 
These methods were then extended by learning-based methods~\cite{bhatnagar2019multi,mir2020learning} that employ differentiable rendering to dress the SMPL~\cite{loper2023smpl} model with a desired garment mesh. 
Despite their effectiveness, such methods rely on costly 3D scanning and the physical presence of human body/clothing, limiting their application in the real world.
Addressing this limitation, M3D-VTON~\cite{zhao2021m3d} proposes a depth-based 3D VTON framework to reconstruct 3D clothed human models from 2D human and garment images, but the results often suffer from explicit warping artifacts.
To improve 3D VTON results, recent methods~\cite{xie2024dreamvton, huang2024tech,huang2024dreamwaltz,zhuang2024dagsm} resort to text-to-image (T2I) diffusion models and employ the Score Distillation Sampling (SDS) loss~\cite{ruiz2022dreambooth} to ensure consistency among different viewing directions.
Specifically, TeCH~\cite{huang2024tech} adapts the generative priors of T2I diffusion model to the specific person and clothes by training descriptive text prompts with DreamBooth~\cite{ruiz2022dreambooth}. DreamWaltz~\cite{huang2024dreamwaltz} leverages Pose ControlNet~\cite{zhang2023adding} to attain clothed human body models. DreamVTON~\cite{xie2024dreamvton} introduces a multi-concept LoRA~\cite{hu2021lora} to personalize the T2I diffusion model, and uses a template-based optimization mechanism that combines with SDS loss to better preserve patterns on the garment. 
Although effective, these methods often produce results lacking in fidelity, as the concepts learned by T2I models are semantic rather than at the pixel level. 
Concurrent to our work, GaussianVTON~\cite{chen2024gaussianvton} proposes an alternative approach by combining Gaussian Splatting~\cite{kerbl20233d} with pre-trained 2D VTON models and formulate it as an editing task.
However, since there are no 2D VTON methods that can generate multi-view images with 3D consistency, the fidelity and 3D consistency of the clothing generated remain problematic.

\vspace{2mm}
\noindent\textbf{Radiance Field-based 3D Human or Scene Editing.}
Recently, radiance field-based editing has attracted significant interest due to its efficient differentiable rendering capabilities, sparking substantial advancements in text-driven 3D editing.
For example, InstructN2N~\cite{haque2023instruct} employ an image-based diffusion model InstructP2P~\cite{brooks2023instructpix2pix} to modify the rendered image by the user's text description with a variant of the score distillation sampling (SDS)~\cite{poole2022dreamfusion} loss. GaussianEditor~\cite{chen2024gaussianeditor} applies Gaussian Splatting~\cite{kerbl20233d} as 3D representation instead of NeRF, adopting Gaussian semantic tracking to track target Gaussian values, significantly improving editing speed and controllability. To enable accurate location and appearance control, subsequent works~\cite{zhuang2023dreameditor, wang2024gaussianeditor} specify the editing region using the attention score or with a segmentation model. TIP-Editor~\cite{zhuang2024tip} proposes a content personalization step dedicated to the reference image based on LoRA, achieving the editing following a hybrid text-image prompt. 
GaussCtrl~\cite{wu2024gaussctrl} leverage depth conditions and attention-based latent code alignment to achieve 3D-aware multi-view consistent editing instead of iteratively editing single views using SDS loss. 
However, these works primarily focus on global appearance modifications within a text-driven pipeline, while our approach emphasizes preserving fine textural details from different viewing directions throughout the editing process.

\section{Preliminary}

\begin{figure*} [ht]
	\begin{center}
		\includegraphics[width=1.0\linewidth]{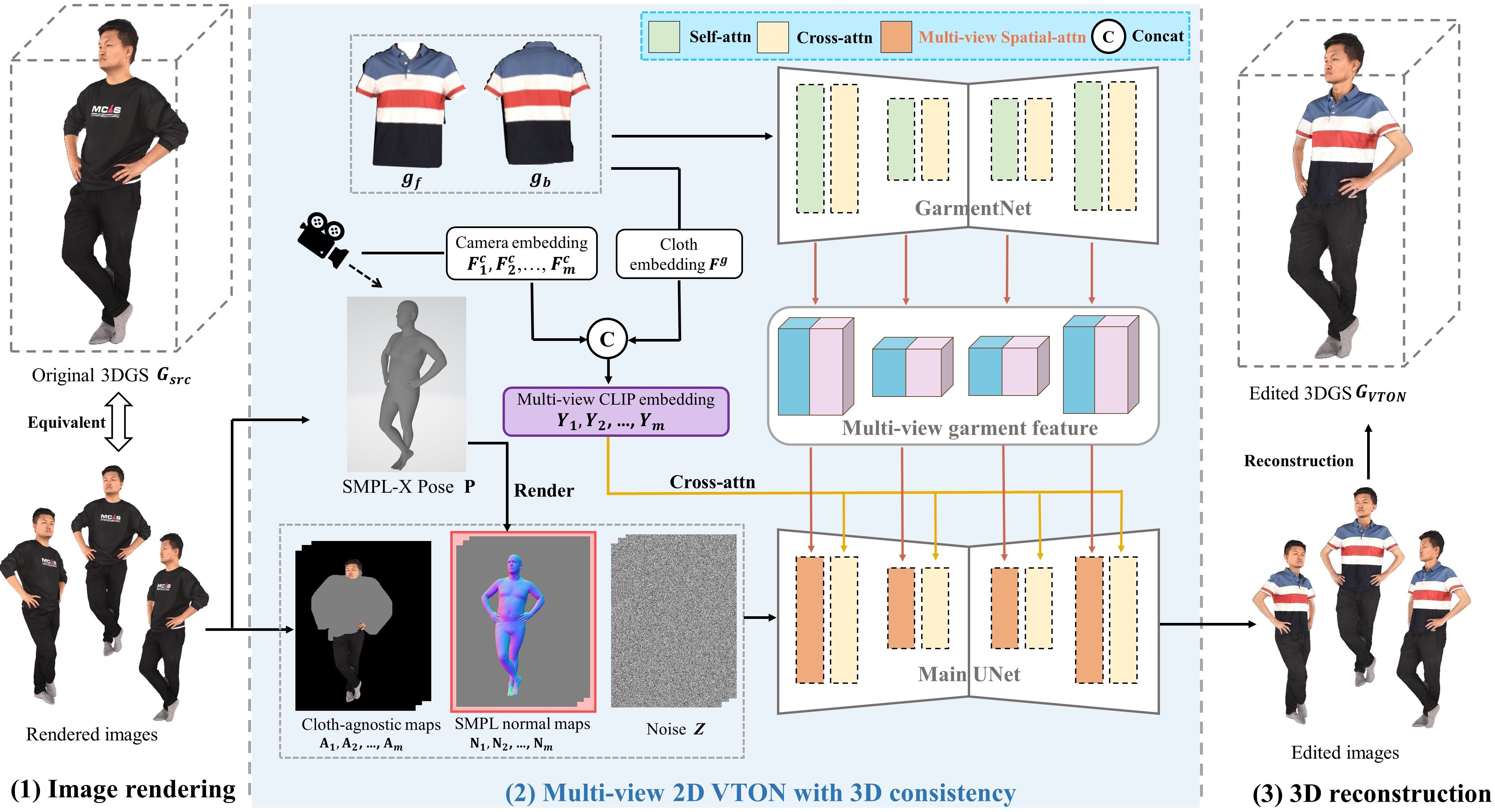} 
	\end{center}
        \vspace{-10pt}
    \caption{\small{\textbf{Overview of VTON 360.} Given an input 3D human model $\mathbf{G_{\rm src}}$ and a pair of garment images ($g_f$, $g_b$), our method 1) renders $\mathbf{G_{\rm src}}$ into multi-view 2D images (left) and 2) edits the rendered multi-view 2D images (middle); 3) reconstructs the edited images into a 3D model $\mathbf{G_{\rm VTON}}$ (right).
    In the crucial step 2), we propose three novel techniques to equip a typical 2D VTON network with the capability to generate 3D-consistent results: \textcolor{red}{1) Pseudo-3D Pose Input}, \textcolor{myorange}{2) Multi-view Spatial Attention}, and \textcolor{mypurple}{3) Multi-view CLIP Embedding}.
    }}
 \label{fig:pipeline}
 \vspace{-15pt}
\end{figure*}

\noindent\textbf{Latent Diffusion Model}.
Latent Diffusion Model~\cite{rombach2022high} is a variant of diffusion models that performs denoising within the latent space of a Variational Autoencoder (VAE)~\cite{kingma2013auto}. During training, given a fixed encoder $\mathcal{E}$, an input image $x$ is transformed into its latent representation $z_0=\mathcal{E}(x)$. A conditional diffusion model $\hat{\boldsymbol{\epsilon}}_{\theta}$, typically implemented with a UNet architecture, is then trained using a weighted denoising score matching objective:
\begin{equation}
\label{eq:diffusion}
    \mathcal{L}_{LDM}=\mathbb{E}_{\mathbf{z},\mathbf{c},\boldsymbol{\epsilon},t}[\boldsymbol{\epsilon} -{\|\hat{\mathbf{\epsilon}}_{\theta}(\mathbf{z_t}; \mathbf{c},t)  \|^2_2}]
\end{equation}
where $\mathbf{z}_t :=\alpha_t \mathbf{x} + \sigma_t \boldsymbol{\epsilon}$ denotes the forward diffusion process at timestep $t$;
$\alpha_t, \sigma_t$ are time-dependent functions defined by the diffusion model formulation;
$\mathbf{c}$ denotes the conditional input and $\boldsymbol{\epsilon} \sim \mathcal{N}(\mathbf{0}, \mathbf{1})$ is Gaussian noise.  
During inference, data samples are generated by initiating from Gaussian noise $\mathbf{z}_{T} \sim \mathcal{N}(\mathbf{0}, \mathbf{1})$ and iteratively refining it using a DDIM~\cite{song2020denoising} sampler.

\section{Method}

Our method leverages the {\it equivalence} between a 3D model and its rendered multi-view 2D images to achieve high-fidelity, any-view 3D VTON. 
Specifically, as Fig.~\ref{fig:pipeline} shows, given an input 3D human model and a garment image, our method 1) renders the 3D model into multi-view 2D images and 2) formulates 3D VTON as a consistent, unified 2D VTON process across these rendered views; 3) By reconstructing the edited images into a 3D model using existing 3D reconstruction methods, we ensure visual coherence and precise garment alignment from any viewing angle.
Among them, the second step is crucial as existing 2D VTON methods~\cite{xu2024ootdiffusion,kim2023stableviton,choi2024improving} lack 3D knowledge, preventing them from generating multi-view images with 3D consistency.

To address this challenge, we propose several novel techniques (Sec.~\ref{sec:image_edit}) that equip a typical 2D VTON network (Sec.~\ref{sec:preliminaries}), which is built on a latent diffusion model~\cite{rombach2022high}, with the capability to generate 3D-consistent results. We use Gaussian Splatting~\cite{kerbl20233d} as our 3D representation.


\begin{figure}
  \centering
   \includegraphics[width=1.0\linewidth]{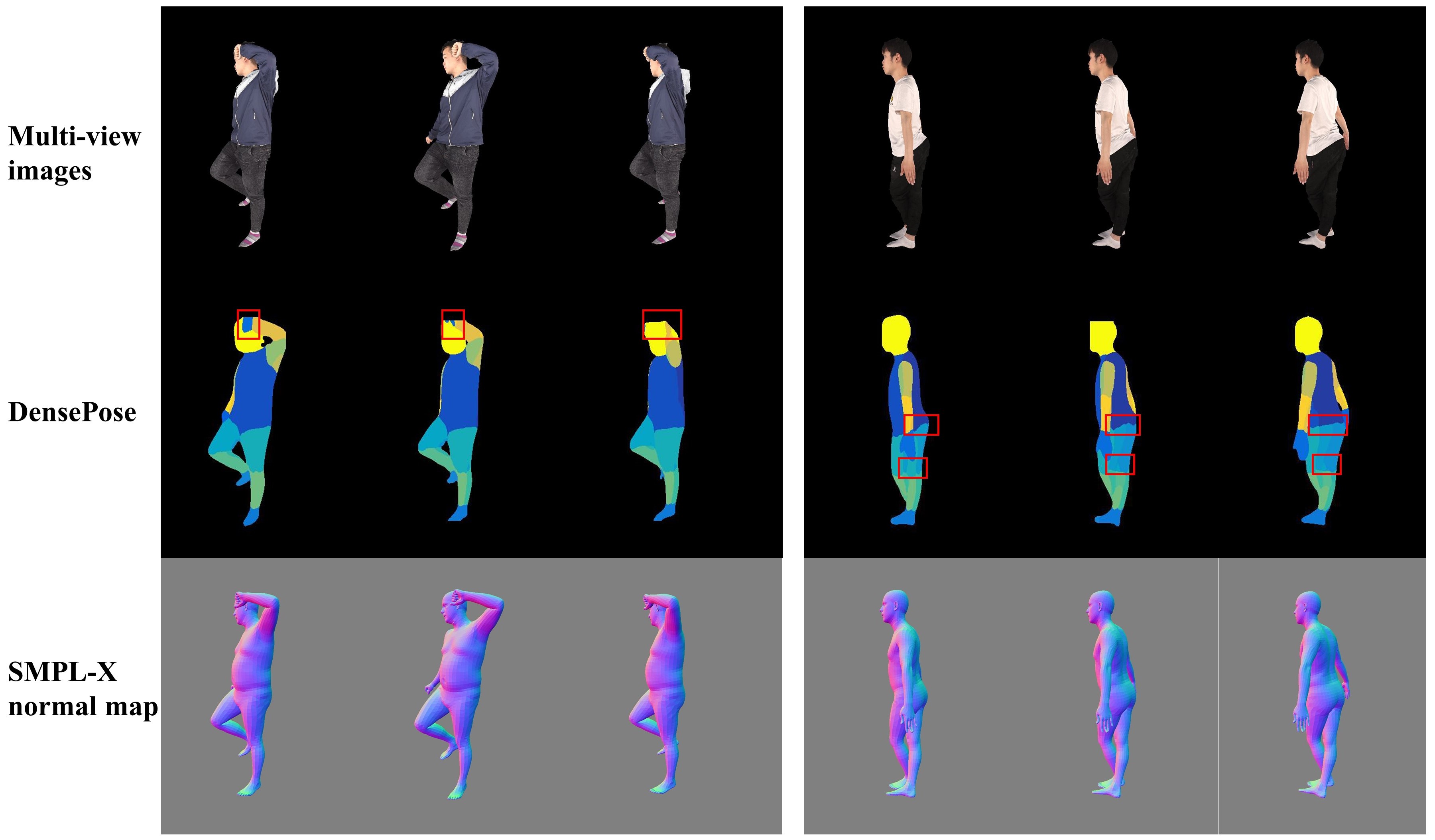}
   \vspace{-15pt}
   \caption{\textbf{DensePose (2D) vs. SMPL-X normal map (pseudo-3D) representations.}
   DensePose applies uniform labels per body part, lacking 3D consistency across views and causing artifacts and temporal inconsistencies (highlighted with red boxes). In contrast, SMPL-X normal maps capture fine surface details, ensuring geometric coherence and stable, realistic shading across views.}
   \label{fig:2D_pose_failure}
   \vspace{-15pt}
\end{figure}

\subsection{Recap of 2D VTON Framework}
\label{sec:preliminaries}

Following previous works~\cite{gou2023taming,xu2024ootdiffusion, kim2023stableviton}, we formulate 2D VTON as an exemplar-based inpainting problem, aiming to fill an input clothing-agnostic image $\mathbf{A}$ with a given garment image $g$, where $\mathbf{A}$ is obtained following the method used in~\cite{xu2024ootdiffusion}. 
As illustrated in Fig.~\ref{fig:pipeline} (middle), the network architecture is based on the latent diffusion model~\cite{rombach2022high} with an encoder $\mathcal{E}$ and comprises two components:
\begin{itemize}
    \item A GarmentNet~\cite{xu2024ootdiffusion, choi2024improving} that extracts features from $\mathcal{E}(g)$.
    \item A Main UNet that inpaints $\mathbf{A}$ according to i) detailed garment features extracted by the GarmentNet; ii) the 2D pose of $\mathbf{A}$ represented by semantic labels using DensePose~\cite{guler2018densepose}; iii) CLIP embeddings of input garment $g$.
    Among them, i) and ii) together with noise are input to the self-attention layers, while iii) is input to the cross-attention layers of the Main UNet.
\end{itemize}
Both the GarmentNet and the Main UNet share the same network architecture.

\subsection{Multi-view 2D VTON with 3D Consistency}
\label{sec:image_edit}

To enable the aforementioned 2D VTON model to generate multi-view and 3D-consistent results, we propose the following novel enhancements to its design:

\noindent\textbf{Multi-view Inputs.} We extend both inputs to the model:
\begin{itemize}
    \item {\it Multi-view Garment Inputs:} We extend the input garment representation from a single image $g$ to paired front and back view images $g_f$, $g_b$, providing comprehensive garment information across all viewing angles.
    Accordingly, we use the encoder $\mathcal{E}$ to encode $g_f, g_b$ into their latent representations $\mathcal{E}(g_f), \mathcal{E}(g_b)$ and feed them into GarmentNet to obtain their multi-layer features $F^l_f$ and $F^l_b$ at layer $l$, respectively.
    \item {\it Multi-view Clothing-agnostic Image Inputs:} Based on the {\it equivalence} between a 3D human model and its rendered multi-view 2D images, we extend the input human body representation from a single, clothing-agnostic image, $\mathbf{A}$, to a set of $m$ multi-view images, denoted as ${\mathbf{A_1}, \mathbf{A_2}, ..., \mathbf{A_m}}$. These images are sampled from random azimuth angles, allowing the 2D VTON model to access comprehensive, multi-view information from the input 3D human model.
\end{itemize}

\noindent\textbf{Pseudo-3D Pose Input.}
As shown in Fig.~\ref{fig:2D_pose_failure}, the 2D DensePose representations~\cite{guler2018densepose} commonly used in state-of-the-art 2D VTON methods~\cite{choi2024improving,kim2023stableviton} assign a uniform semantic label to all pixels within each body part (\eg, thigh), inherently lack 3D geometric consistency across multiple views, and often introduce artifacts and temporal inconsistencies.
To address these limitations, we propose a novel pseudo-3D pose representation: the normal maps $\mathbf{N}$ derived from the SMPL-X~\cite{pavlakos2019expressive} model of the input body. These normal maps capture fine-grained surface orientation details, providing a more consistent representation across views by preserving geometric structure in the 3D space. Furthermore, they enable smoother, temporally stable transitions and realistic shading effects across multi-view scenarios. 
In practice, we employ a lightweight PoseEncoder $\mathcal{E'}$~\cite{hu2023animate} and feed $\mathcal{E'}(\mathbf{N})$ into the Main UNet. We obtain the SMPL-X model from the multi-view images of the input body using EasyMoCap~\cite{easymocap}.

Accordingly, we concatenate three components as the enhanced input to the Main UNet: i) a noise latent $z_t$; ii) the encoded pseudo-3D poses $\mathcal{E'}(\mathbf{N_1}), \mathcal{E'}(\mathbf{N_2}), ..., \mathcal{E'}(\mathbf{N_m})$; and iii) the encoded multi-view clothing-agnostic images $\mathcal{E}(\mathbf{A_1}), \mathcal{E}(\mathbf{A_2}), ..., \mathcal{E}(\mathbf{A_m})$.
Let $F^l_1, F^l_2, ..., F^l_m$ be the feature representations at layer $l$ of the Main UNet, and recall the garment features $F^l_f$ and $F^l_b$ defined above, we enhance the self-attention layers of the Main UNet as:

\noindent\textbf{Multi-view Spatial Attention.}
To cope with the aforementioned multi-view input features and ensure their consistency, we draw insights from the {\it temporal} attention layer commonly used in video generation and editing~\cite{zhang2023controlvideo,wu2023tune} and extend it to our multi-view {\it spatial} attention layer, denoted as $\rm MVAttention$. The key distinction of our $\rm MVAttention$ is that its input multi-view features $F^l_1, F^l_2, ..., F^l_m$ are from images captured from non-uniform spatial intervals, with the viewing angles varying randomly.
Consequently, features from similar views exhibit a higher correlation, while those from distinct views are largely independent.
To model this relationship, we construct a ``correlation'' matrix $C$ based on the angular disparity obtained from camera rotation matrices of the input multi-view images, and define our $\rm MVAttention$ as follows:
\begin{equation}
\begin{aligned}
    \mathbf{F^l}&=[F^l_1\oplus F^l_2 \ldots \oplus F^l_m],\ \mathbf{\hat{F}^l}=[\mathbf{F^l}\oplus F^l_f\oplus F^l_b]\\
    Q&=W^Q \ \mathbf{F^l}, \  K=W^K \ \mathbf{\hat{F}^l},\  V=W^V \ \mathbf{\hat{F}^l}\\
    A_i&=\text{softmax}(\frac{Q_i\times (C_i\cdot K^T)}{\sqrt{d}}), \ \mathbf{H^l_i}=A_i\times V_i
    \label{eq:full_attn}
\end{aligned}
\end{equation}
where $i\in \{1,2,...,m\}$ denotes $i$-th view; the Query $Q$ comes directly from $\mathbf{F^l}$ and the concatenation of $[\mathbf{F^l}, F^l_f, F^l_b]$ serves as the key $K$ and the value $V$; $\oplus$ indicates matrix concatenation along the token axis; $d$ denotes the dimension; $W^Q$, $W^K$, $W^V$ represent the linear transformation matrices; we omitted the $l$ of the attention matrices and parameters for simplicity; $C\in \mathbb{R}^{m\times m}$, $C_i$ represents $i$-th row in $C$, and its ``correlation'' value between $i$-th and $j$-th features is 
$C_{ij}$:
\begin{equation}
     C_{ij}=((\text{trace}(R_i^T R_j) - 1) / 2+1)/2
\end{equation}
where $R_i$ and $R_j$ are the extrinsic rotation matrices of the corresponding camera views, $(\text{trace}(R_i^T R_j) - 1) / 2$ is the cosine value of the angle between these camera views. 

\noindent\textbf{Multi-view CLIP Embedding.}
Camera viewpoints can serve as an effective condition signal to enhance 3D consistency in video content generation~\cite{xu2024camco}. Building on this insight, we incorporate camera condition within our try-on network by encoding camera parameters as an additional token, enabling the generation of more consistent multi-view images.
Specifically, we define a world coordinate system in which the camera faces the subject directly. For each input image (view) $\mathbf{A_i}$, $1\leq i \leq m$, we extract the rotation matrix from the camera's corresponding extrinsic matrix. This rotation matrix is then reshaped into a 9-dimensional tensor $\mathbf{r_i}$, which undergoes positional encoding to effectively integrate the camera parameters into the feature representation $F^c_i$.
\begin{equation}
\begin{split}
    F^c_i=(\sin(2^0\pi\mathbf{r_i}), \cos(2^0\pi\mathbf{r_i}),
..., \\ \sin(2^{L-1}\pi\mathbf{r_i}), \cos(2^{L-1}\pi\mathbf{r_i}))
\end{split}
\end{equation}
where $L$ is the length of positional embedding. 
We then project $F^c_i$ to match the dimensionality of the garment CLIP image embedding $F^g$ via an MLP and concatenate them along the token axis to form $Y_i$. This combined representation, $Y_i$, is subsequently used in the key $K_{\rm x}$ and value $V_{\rm x}$ of the cross-attention layers of the Main UNet:
\begin{equation}
\begin{aligned}
    Y_i=F^g \oplus &{\rm MLP}(F^c_i) \\
    Q_{\rm x}=W_{\rm x}^Q \ \mathbf{H^l_i}, \  K_{\rm x}=&W_{\rm x}^K \ Y_i,\  V_{\rm x}=W_{\rm x}^V \  Y_i\\
    F^{(l+1)}_i=\text{softmax}&(\frac{Q_{\rm x}K_{\rm x}^T}{\sqrt{d_{\rm x}}})V_{\rm x},
    \label{eq:cross_attn}
\end{aligned}
\end{equation}
where $\mathbf{H^l}$ is the output of the $\rm MVAttention$ of the $l$-th layer; we omitted the $l$ of the cross attention matrices and parameters for simplicity.

\noindent \textbf{Training.}
Our enhanced multi-view 2D VTON network can be trained by minimizing the following latent diffusion model loss function:
\begin{equation}\label{eq:ldm_loss}
    \mathcal{L}_{\rm ldm} = \mathbb{E}_{z_t,\eta, \psi, \epsilon,t}\left[\lVert\epsilon - \epsilon_{\theta}(z_t, t,\eta,\psi, \mathbf{\zeta})) \rVert_2^2\right],
\end{equation}
where \begin{small}$\eta= [\mathcal{E}(g_f);\mathcal{E}(g_b);\mathcal{E}(\mathbf{N_i})^{m}_{i=1}]$\end{small} represents the input latent garment images and latent normal maps; \begin{small}$\mathbf{\zeta}=[\mathcal{E}'(\mathbf{A_i})^m_{i=1}]$\end{small} denotes the input latent clothing-agnoistic images; $\psi=\mathbf{Y}$ is the proposed multi-view CLIP embedding.

\begin{figure}
  \centering
   \includegraphics[width=1.0\linewidth]{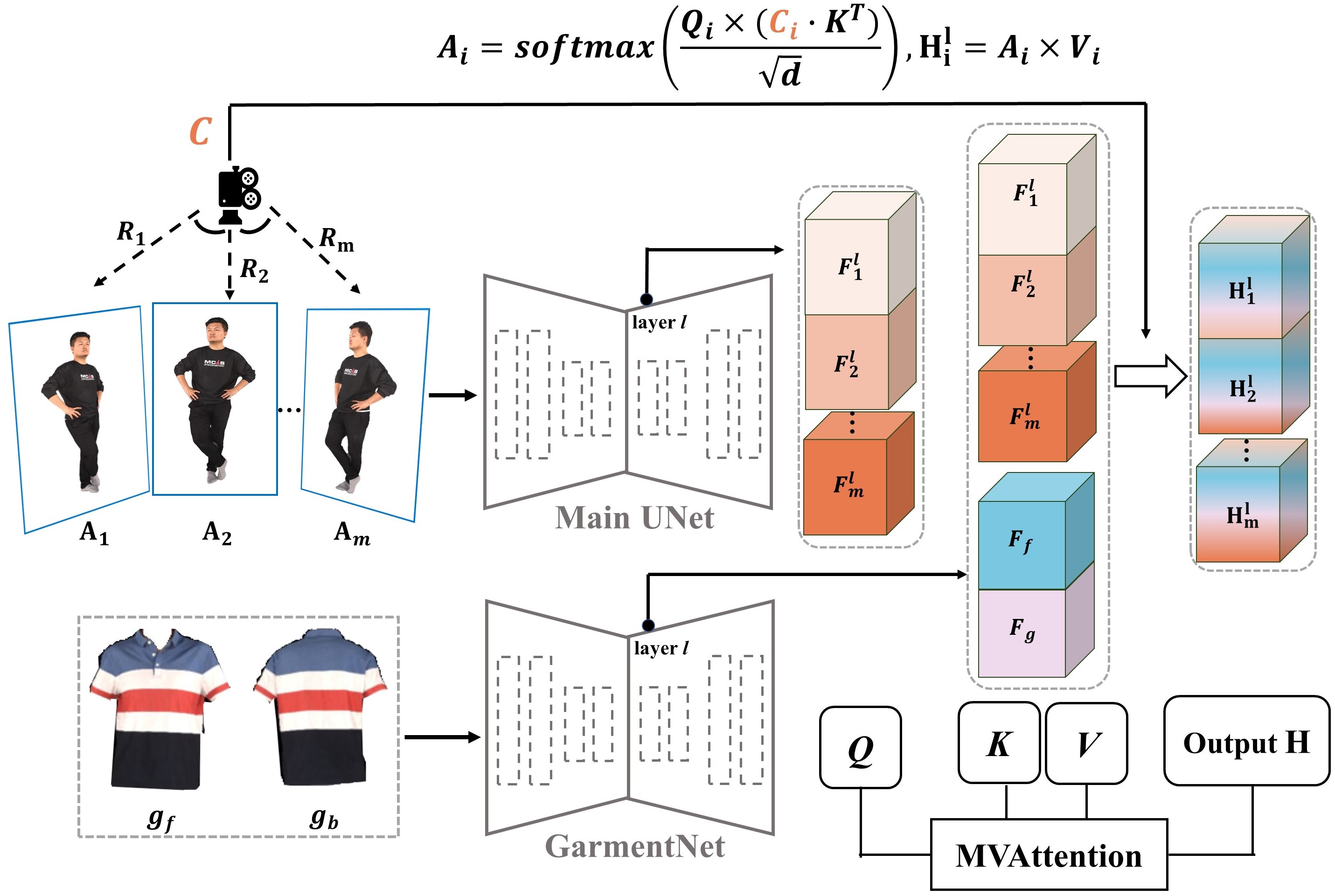}
      \caption{\textbf{Illustration of the proposed Multi-view Spatial Attention.} Query (Q): multi-view features $\mathbf{F^l}$; Key (K) and Value (V): concatenation of $\mathbf{F^l}$ and garment features $F^l_f$ and $F^l_b$. The attention score between viewpoints $i$ and $j$ is modulated by a weight $C_{ij}$, determined by the cosine of the angle between them.}
   \label{fig:attn}
   \vspace{-15pt}
\end{figure}

\section{Experiments}

\subsection{Experimental Setup}
\noindent\textbf{Datasets.}
We conduct experiments on two public datasets: Thuman2.0~\cite{yu2021function4d} and MVHumanNet~\cite{xiong2024mvhumannet}. Thuman2.0 comprises 526 reconstructed clothed human scans, from which we render multi-view input images. Of these samples, 426 are used for training, while the remaining 100 are set aside for testing. 
To further evaluate the effectiveness and robustness of our method, we also perform experiments on MVHumanNet, a large-scale dataset of multi-view human images that encompasses a diverse range of subjects, daily outfits and motion sequences. The images in MVHumanNet are captured using a multi-view system with either 48 or 24 cameras. We use 4,990 subjects from this dataset, allocating 4,790 to training and 200 for tests.
For each subject, we randomly select two frames of multi-view images from its entire motion sequence. 
While MVHumanNet provides multi-view images directly for editing and reconstruction, we render uniformly distributed views around each human subject in Thuman2.0 to ensure consistent input.
\begin{table*}[htp]
\centering
    \resizebox{\textwidth}{!}{
    \begin{tabular}{lcccccccccc}
    \hline
    \multirow{2}*{Method}& \multirow{7}{*}{} & \multicolumn{4}{c}{Thuman2.0~\cite{yu2021function4d}}   & \multirow{7}{*}{} & \multicolumn{4}{c}{MVHumanNet~\cite{xiong2024mvhumannet}} \\  \cline{3-6} \cline{8-11}
    &  & CLIP$_{cons} \uparrow$ & DINO$_{sim} \uparrow$ &  Vote$_{quality}$ & Vote$_{align}$&  & CLIP$_{cons} \uparrow$ & DINO$_{sim} \uparrow$ &  Vote$_{quality}$ & Vote$_{align}$ \\ \cline{1-1}\cline{3-6} \cline{8-11}
    DreamWaltz~\cite{huang2024dreamwaltz}& & 0.887 & 0.556 & 0.46\% & 1.54\% & & 0.935 & 0.495 & 0.46\% & 0.46\%\\
    TIP-Editor~\cite{zhuang2024tip}& & \textbf{0.939} & 0.569 & 0.92\% & 0.62\% &  & \textbf{0.948} & 0.512  & 2.15\% & 1.38\%\\
    GaussCtrl~\cite{wu2024gaussctrl}& & 0.931 & 0.577  & 1.08\% & 1.38\% & & 0.938 & 0.521  & 1.69\% & 1.23\%\\
    \midrule
    Ours & & 0.923 & \textbf{0.633} & \textbf{97.54}\% & \textbf{96.46}\% & & 0.933 & \textbf{0.623}  & \textbf{95.69}\% & \textbf{96.92}\%\\
    \bottomrule
    \end{tabular}
    }
    \vspace{-5pt}
    \caption{\textbf{Quantitative comparisons.} CLIP$_{cons}$ denotes the CLIP Direction Consistency Score. DINO$_{sim}$ is the DINO similarity.}
    \label{table:comparison}
    \vspace{-10pt}
\end{table*}

\begin{figure*} [ht]
	\begin{center}
		\includegraphics[width=.9\linewidth]{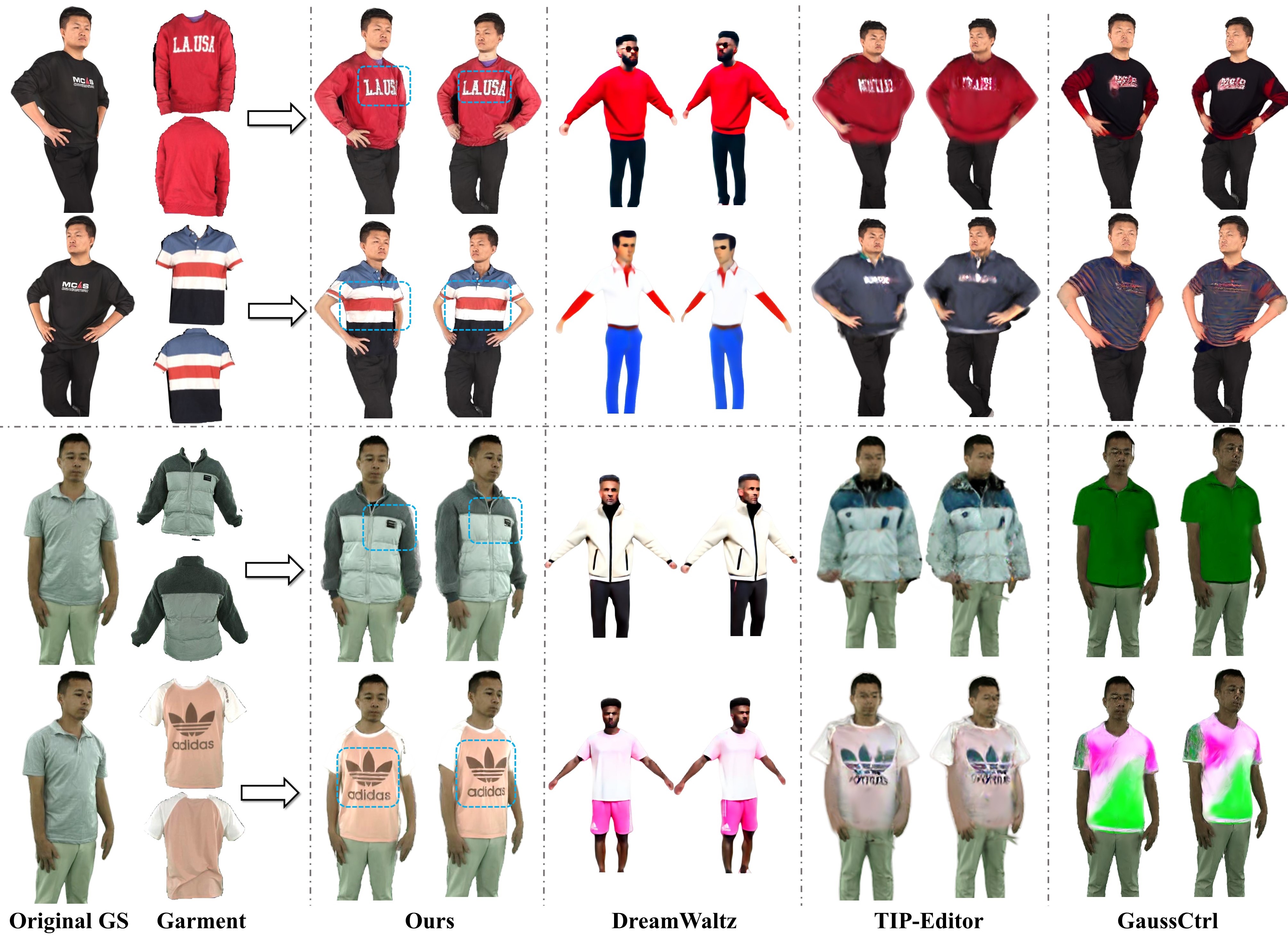} 
	\end{center}
        \vspace{-10pt}
	\caption{\small{\textbf{Qualitative comparison.} The first two rows show the results on Thuman2.0 dataset while the last two rows show the results on MVHumaNet dataset. Our method achieves good texture preservation (highlighted by the blue boxes), while three baseline methods mostly fail. }}\label{fig:qualitative_comparison}
 \vspace{-12pt}
\end{figure*}

\vspace{1mm}
\noindent\textbf{Baselines.}
We primarily compare our method with three existing methods: DreamWaltz~\cite{huang2024dreamwaltz}, GaussCtrl~\cite{wu2024gaussctrl}, and TIP-Editor~\cite{zhuang2024tip}. DreamWaltz is a method designed for directly generating 3D human bodies based on textual descriptions, while GaussCtrl and TIP-Editor are two radiance-based editing methods.
GaussCtrl is based on Stable Diffusion, using a description-like prompt to edit the scene. TIP-Editor accepts both text and image prompts. We configure it by specifying the human body as the editing region and the desired garment as the image prompt. We use ChatGPT to generate the text prompts corresponding to the clothing images. 

\vspace{1mm}
\noindent\textbf{Evaluation Metrics.}
For quantitative evaluation, we assess garment-to-person alignment between the edited person and the reference image. Following~\cite{zhuang2024tip}, we calculate the average DINO similarity~\cite{oquab2023dinov2} between the reference image and the rendered multi-view images of the edited 3D scene. Additionally, to evaluate multi-view consistency, we compute the CLIP Directional Consistency Score as outlined in~\cite{haque2023instruct}.
Given the large scale of experiments (repeated 3DGS reconstruction), we selected a subset of examples from the dataset for metric evaluation. Specifically, from the test sets of Thuman and MVHumanNet, we randomly sampled 10 human scans each, performing virtual try-on with 6 randomly chosen garments per human scan.

We further conducted a user study involving 50 participants who rated the results of our method and three baseline methods based on two criteria: overall ``Quality'' and ``Alignment'' with the reference image.
Each evaluation consisted of two questions: (1) Which method produces the highest quality of the edited 3D human? and (2) Which method achieves the most consistent alignment with the target clothing? Participants viewed the VTON results as rotating randomized video sequences.



\vspace{1mm}
\noindent\textbf{Implementation Details.}
During pre-processing, we crop the multi-view images to the bounding box around the person and resize them to a resolution of $768 \times 576$. The front view and the back view of garment images are obtained from the corresponding clothed human images. After editing, we pad the images back to their original size. The data processing pipeline is the same for both Thuman2.0 and MVHumanNet datasets.

The Main UNet and the GarmentNet are initialized by the Stable Diffusion V1.5~\cite{rombach2022high}. The training process is divided into two stages. In the first stage, each view is trained independently, during which we establish the feature extraction capabilities of both the PoseEncoder and GarmentNet, as well as the generative capability of the Main UNet. The second stage involves multi-view training, where we randomly select $M$ views for each human subject. This stage is focused on training the proposed $\rm MVAttention$ module to enhance the network’s ability to maintain consistency across views. Due to memory constraints, we set $M=8$ for the training phase. During the testing phase, we uniformly sampled 32 views from a 360-degree rotation around the subject. The editing of these 32 views is conducted in two batches, with each batch processing $M=16$ views. 



\subsection{Comparisons with State-of-the-Art Methods}
\noindent\textbf{Qualitative Evaluation.}
Fig.~\ref{fig:qualitative_comparison} shows visual comparisons between our method and the baselines. 
DreamWaltz~\cite{huang2024dreamwaltz} regenerates 3D clothed humans from text prompts but struggles to accurately retain both body and clothing characteristics. 
GaussCtrl~\cite{wu2024gaussctrl}, lacking support for image prompts, fails to maintain detailed clothing textures. 
While Tip-Editor~\cite{zhuang2024tip} leverages Lora~\cite{hu2021lora} for personalization, it encounters difficulties in consistently mapping clothing inputs from two views into the 3D human because the personalized concept are semantic in 2D space. In contrast, our method effectively preserves intricate clothing details, such as text, stripes, and logos.


\vspace{1mm}
\noindent\textbf{Quantitative Evaluation.}
Tab.~\ref{table:comparison} shows the results for the CLIP Directional Consistency Score and DINO similarity on Thuman2.0 and MVHumanNet datasets. Our approach surpasses other methodes on DINO$_{sim}$, clearly illustrating the superiority of our method in terms of garment texture preservation. While our results on CLIP$_{cons}$ are comparable to those of other methods, it is important to note that these methods incorporate the SDS loss, which to some extent smooths the representation of humans in 3D space. Additionally, the "flatter" textures of other methods could also result in artificially higher consistency scores. Furthermore, user studies have shown that our method significantly exceeds baselines in terms of edited 3D human quality and the alignment of clothing details.


\subsection{Visual Results using E-commerce Garment}
\begin{figure}
  \centering
   \includegraphics[width=1.0\linewidth]{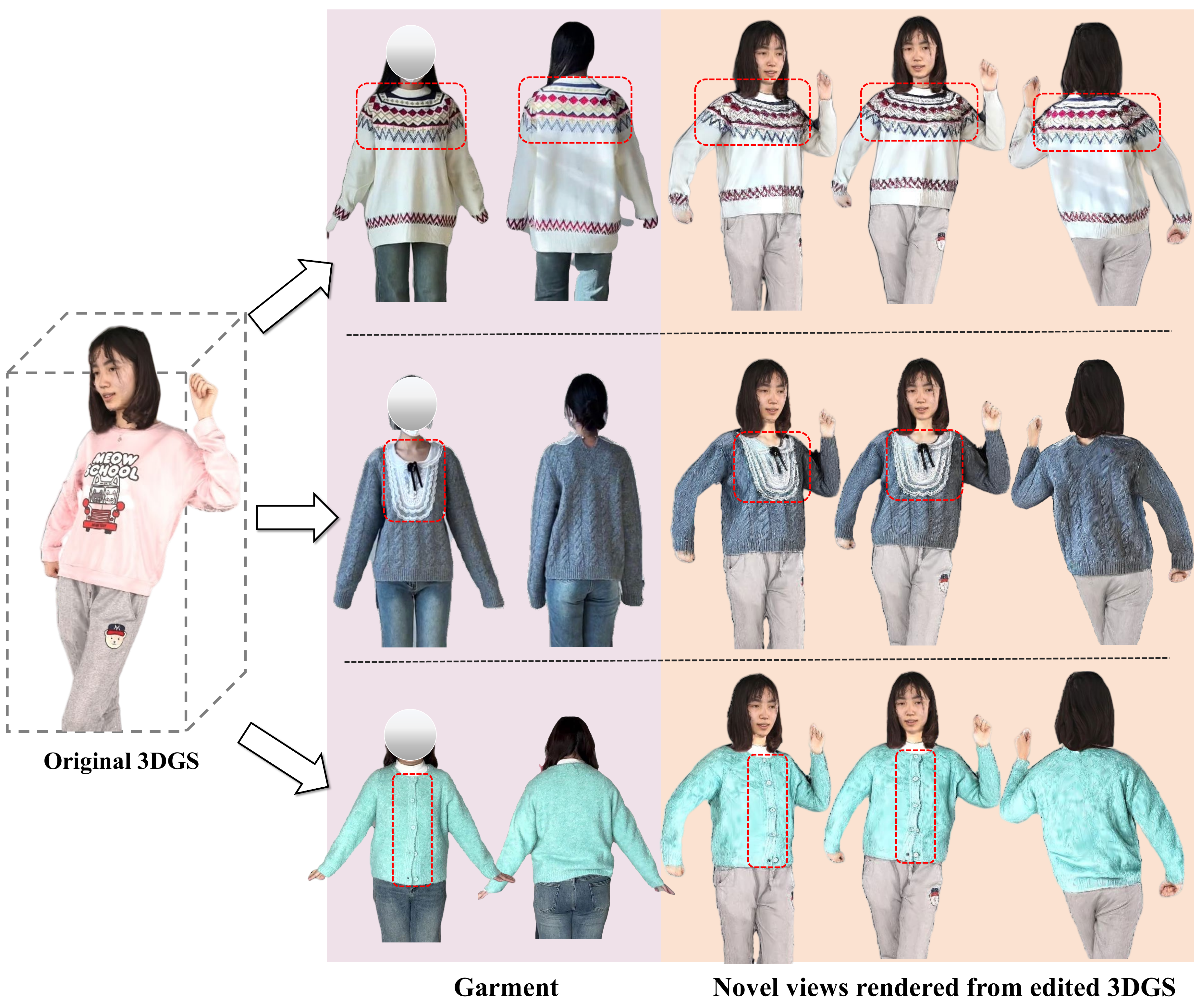}
   \vspace{-5pt}
    \caption{\textbf{Generalization to e-commerce garments (the MVG dataset).}
    Our method, trained on the THuman2.0 dataset, shows strong generalizability when applied to e-commerce garments. For clarity in visualization, we display garment images on human models; however, in the actual VTON process, the garments are segmented from the models using parse maps.}
   \label{fig:mvg}
   \vspace{-15pt}
\end{figure}

Fig.~\ref{fig:mvg} showcases VTON results using garments from the MVG dataset~\cite{wang2024mv}, whose images are from e-commerce platforms like YOOX NET-A-PORTER, Taobao, and TikTok\footnote{https://net-a-porter.com, www.taobao.com, www.douyin.com}, and a model trained on the Thuman2.0 dataset~\cite{yu2021function4d}. 
The results demonstrate that our method effectively preserves intricate garment details and textures. For instance, it accurately maintains the stripe patterns in the first row, the cute tie in the second row, and the buttons in the third row, highlighting the robustness of our approach in handling diverse and realistic clothing items.



\begin{figure}
  \centering
   \includegraphics[width=0.9\linewidth]{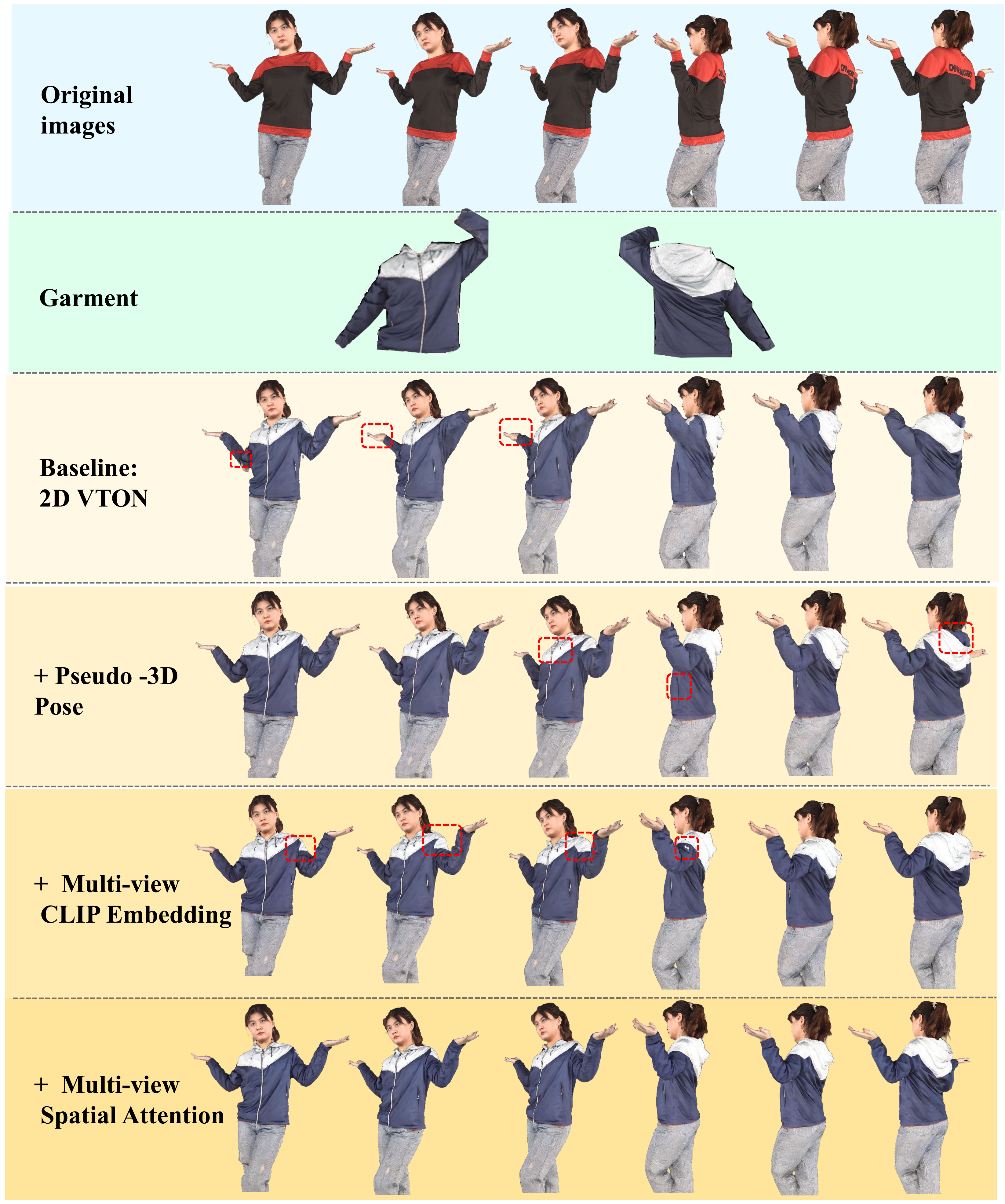}
       \vspace{-5pt}
       \caption{\textbf{Visualization of the impact of the three proposed techniques on multi-view consistent editing.} The red boxes highlight the artifacts. Starting from the 2D VTON baseline, the pseudo-3D pose improves limb generation quality, multi-view CLIP embedding enhances detail across different viewing directions, and finally, $\rm MVAttention$ further strengthens consistency in the generated images.}
   \label{fig:ablation}
   \vspace{-10pt}
\end{figure}

\subsection{Ablation Study}
We conduct an ablation study on Thuman2.0 dataset in Tab.~\ref{table:ablation} and Fig.~\ref{fig:ablation} to evaluate the impact of our three proposed modules in enhancing a typical 2D VTON network with 3D-consistent generation capabilities. 
Starting with the 2D VTON baseline~\cite{xu2024ootdiffusion} using DensePose, we progressively replace DensePose with our pseudo-3D pose, incorporate multi-view CLIP embeddings, and ultimately integrate $\rm MVAttention$ in the final configuration. 
Results in Tab.~\ref{table:ablation} indicate that each module contributes to metric improvements.
Fig.~\ref{fig:ablation} visualizes an example of multi-view image editing. The incorporation of pseudo-3D pose substantially improves limb generation compared to the 2D VTON baseline. 
Comparing rows 4 and 5, prior to the integration of multi-view CLIP embedding, the model captures limited spatial information, resulting in detail loss at specific angles (columns 3, 4, and 6). Finally, the proposed $\rm MVAttention$ achieves a more coherent generation across views.



\begin{table}[h]
\setlength\tabcolsep{2pt}%
    \centering
    \begin{tabular}{l|c|c}
    \toprule
    Methods & CLIP$_{cons}$ $\uparrow$ & DINO$_{sim}$ $\uparrow$  \\
    \midrule
    2D-VTON & 0.892 & 0.609  \\
    +~Pseudo-3D Pose & 0.910 & 0.626 \\
    +~Multi-view CLIP Embedding & 0.913 & 0.631 \\
    +~Multi-view Spatial Attention & \textbf{0.923} & \textbf{0.633}\\
    \bottomrule
    \end{tabular}
\vspace{-5pt}
\caption{\textbf{Ablation studies.} We ablate the impact of the three proposed techniques on Thuman2.0 dataset.}
\vspace{-12pt}
\label{table:ablation}
\end{table}

\section{Conclusions}

In this work, we proposed VTON 360, a novel 3D Virtual Try-On (VTON) method that achieves high-fidelity VTON with the ability to render clothing from arbitrary viewing directions. 
Our method features a novel formulation of 3D VTON as an extension of 2D VTON that ensures 3D consistent results across multiple views.
To bridge the gap between 2D VTON models and 3D consistency requirements, we introduce several key innovations, including multi-view inputs, pseudo-3D pose representation, multi-view spatial attention, and multi-view CLIP embedding. Extensive experiments demonstrate the effectiveness of our approach, significantly outperforming prior 3D VTON techniques in both fidelity and any-view rendering.

\section*{Acknowledgement}
This work is supported in part by the National Key R\&D Program of China under Grant No.2024YFB3908503, in part by the National Natural Science Foundation of China under Grant NO.~62322608 and in part by the CCF-Kuaishou Large Model Explorer Fund (NO.~CCF-KuaiShou 2024007).

{
    \small
    \bibliographystyle{ieeenat_fullname}
    \bibliography{main}

\begin{thebibliography}{63}
\providecommand{\natexlab}[1]{#1}
\providecommand{\url}[1]{\texttt{#1}}
\expandafter\ifx\csname urlstyle\endcsname\relax
  \providecommand{\doi}[1]{doi: #1}\else
  \providecommand{\doi}{doi: \begingroup \urlstyle{rm}\Url}\fi

\bibitem[eas(2021)]{easymocap}
Easymocap - make human motion capture easier.
\newblock Github, 2021.

\bibitem[Bai et~al.(2022)Bai, Zhou, Li, Zhou, and Yang]{bai2022single}
Shuai Bai, Huiling Zhou, Zhikang Li, Chang Zhou, and Hongxia Yang.
\newblock Single stage virtual try-on via deformable attention flows.
\newblock In \emph{European Conference on Computer Vision}, pages 409--425. Springer, 2022.

\bibitem[Bhatnagar et~al.(2019)Bhatnagar, Tiwari, Theobalt, and Pons-Moll]{bhatnagar2019multi}
Bharat~Lal Bhatnagar, Garvita Tiwari, Christian Theobalt, and Gerard Pons-Moll.
\newblock Multi-garment net: Learning to dress 3d people from images.
\newblock In \emph{Proceedings of the IEEE/CVF international conference on computer vision}, pages 5420--5430, 2019.

\bibitem[Bridson et~al.(2002)Bridson, Fedkiw, and Anderson]{bridson2002robust}
Robert Bridson, Ronald Fedkiw, and John Anderson.
\newblock Robust treatment of collisions, contact and friction for cloth animation.
\newblock In \emph{Proceedings of the 29th annual conference on Computer graphics and interactive techniques}, pages 594--603, 2002.

\bibitem[Brooks et~al.(2023)Brooks, Holynski, and Efros]{brooks2023instructpix2pix}
Tim Brooks, Aleksander Holynski, and Alexei~A Efros.
\newblock Instructpix2pix: Learning to follow image editing instructions.
\newblock In \emph{Proceedings of the IEEE/CVF Conference on Computer Vision and Pattern Recognition}, pages 18392--18402, 2023.

\bibitem[Chen et~al.(2024{\natexlab{a}})Chen, Huang, Huang, Ge, and Shao]{chen2024gaussianvton}
Haodong Chen, Yongle Huang, Haojian Huang, Xiangsheng Ge, and Dian Shao.
\newblock Gaussianvton: 3d human virtual try-on via multi-stage gaussian splatting editing with image prompting.
\newblock \emph{arXiv preprint arXiv:2405.07472}, 2024{\natexlab{a}}.

\bibitem[Chen et~al.(2024{\natexlab{b}})Chen, Chen, Zhang, Wang, Yang, Wang, Cai, Yang, Liu, and Lin]{chen2024gaussianeditor}
Yiwen Chen, Zilong Chen, Chi Zhang, Feng Wang, Xiaofeng Yang, Yikai Wang, Zhongang Cai, Lei Yang, Huaping Liu, and Guosheng Lin.
\newblock Gaussianeditor: Swift and controllable 3d editing with gaussian splatting.
\newblock In \emph{Proceedings of the IEEE/CVF Conference on Computer Vision and Pattern Recognition}, pages 21476--21485, 2024{\natexlab{b}}.

\bibitem[Choi et~al.(2021)Choi, Park, Lee, and Choo]{choi2021viton}
Seunghwan Choi, Sunghyun Park, Minsoo Lee, and Jaegul Choo.
\newblock Viton-hd: High-resolution virtual try-on via misalignment-aware normalization.
\newblock In \emph{Proceedings of the IEEE/CVF Conference on Computer Vision and Pattern Recognition}, pages 14131--14140, 2021.

\bibitem[Choi et~al.(2024)Choi, Kwak, Lee, Choi, and Shin]{choi2024improving}
Yisol Choi, Sangkyung Kwak, Kyungmin Lee, Hyungwon Choi, and Jinwoo Shin.
\newblock Improving diffusion models for virtual try-on.
\newblock \emph{arXiv preprint arXiv:2403.05139}, 2024.

\bibitem[Dong et~al.(2022)Dong, Zhao, Xie, Zhang, Du, Zheng, Long, Liang, and Yang]{dong2022dressing}
Xin Dong, Fuwei Zhao, Zhenyu Xie, Xijin Zhang, Daniel~K Du, Min Zheng, Xiang Long, Xiaodan Liang, and Jianchao Yang.
\newblock Dressing in the wild by watching dance videos.
\newblock In \emph{Proceedings of the IEEE/CVF Conference on Computer Vision and Pattern Recognition}, pages 3480--3489, 2022.

\bibitem[Ge et~al.(2021)Ge, Song, Zhang, Ge, Liu, and Luo]{ge2021parser}
Yuying Ge, Yibing Song, Ruimao Zhang, Chongjian Ge, Wei Liu, and Ping Luo.
\newblock Parser-free virtual try-on via distilling appearance flows.
\newblock In \emph{Proceedings of the IEEE/CVF conference on computer vision and pattern recognition}, pages 8485--8493, 2021.

\bibitem[Gou et~al.(2023)Gou, Sun, Zhang, Si, Qian, and Zhang]{gou2023taming}
Junhong Gou, Siyu Sun, Jianfu Zhang, Jianlou Si, Chen Qian, and Liqing Zhang.
\newblock Taming the power of diffusion models for high-quality virtual try-on with appearance flow.
\newblock \emph{arXiv preprint arXiv:2308.06101}, 2023.

\bibitem[Guan et~al.(2012)Guan, Reiss, Hirshberg, Weiss, and Black]{guan2012drape}
Peng Guan, Loretta Reiss, David~A Hirshberg, Alexander Weiss, and Michael~J Black.
\newblock Drape: Dressing any person.
\newblock \emph{ACM Transactions on Graphics (ToG)}, 31\penalty0 (4):\penalty0 1--10, 2012.

\bibitem[G{\"u}ler et~al.(2018)G{\"u}ler, Neverova, and Kokkinos]{guler2018densepose}
R{\i}za~Alp G{\"u}ler, Natalia Neverova, and Iasonas Kokkinos.
\newblock Densepose: Dense human pose estimation in the wild.
\newblock In \emph{Proceedings of the IEEE conference on computer vision and pattern recognition}, pages 7297--7306, 2018.

\bibitem[Hahn et~al.(2014)Hahn, Thomaszewski, Coros, Sumner, Cole, Meyer, DeRose, and Gross]{hahn2014subspace}
Fabian Hahn, Bernhard Thomaszewski, Stelian Coros, Robert~W Sumner, Forrester Cole, Mark Meyer, Tony DeRose, and Markus Gross.
\newblock Subspace clothing simulation using adaptive bases.
\newblock \emph{ACM Transactions on Graphics (TOG)}, 33\penalty0 (4):\penalty0 1--9, 2014.

\bibitem[Han et~al.(2018)Han, Wu, Wu, Yu, and Davis]{han2018viton}
Xintong Han, Zuxuan Wu, Zhe Wu, Ruichi Yu, and Larry~S Davis.
\newblock Viton: An image-based virtual try-on network.
\newblock In \emph{Proceedings of the IEEE conference on computer vision and pattern recognition}, pages 7543--7552, 2018.

\bibitem[Haque et~al.(2023)Haque, Tancik, Efros, Holynski, and Kanazawa]{haque2023instruct}
Ayaan Haque, Matthew Tancik, Alexei~A Efros, Aleksander Holynski, and Angjoo Kanazawa.
\newblock Instruct-nerf2nerf: Editing 3d scenes with instructions.
\newblock In \emph{Proceedings of the IEEE/CVF International Conference on Computer Vision}, pages 19740--19750, 2023.

\bibitem[He et~al.(2022)He, Song, and Xiang]{He_2022_CVPR}
Sen He, Yi-Zhe Song, and Tao Xiang.
\newblock Style-based global appearance flow for virtual try-on.
\newblock In \emph{Proceedings of the IEEE/CVF Conference on Computer Vision and Pattern Recognition (CVPR)}, pages 3470--3479, 2022.

\bibitem[He et~al.(2024)He, Chen, Wang, Li, Torr, and Lin]{he2024wildvidfit}
Zijian He, Peixin Chen, Guangrun Wang, Guanbin Li, Philip~HS Torr, and Liang Lin.
\newblock Wildvidfit: Video virtual try-on in the wild via image-based controlled diffusion models.
\newblock In \emph{European Conference on Computer Vision}, pages 123--139. Springer, 2024.

\bibitem[Ho et~al.(2020)Ho, Jain, and Abbeel]{ho2020denoising}
Jonathan Ho, Ajay Jain, and Pieter Abbeel.
\newblock Denoising diffusion probabilistic models.
\newblock \emph{Advances in Neural Information Processing Systems}, 33:\penalty0 6840--6851, 2020.

\bibitem[Hu et~al.(2021)Hu, Shen, Wallis, Allen-Zhu, Li, Wang, Wang, and Chen]{hu2021lora}
Edward~J Hu, Yelong Shen, Phillip Wallis, Zeyuan Allen-Zhu, Yuanzhi Li, Shean Wang, Lu Wang, and Weizhu Chen.
\newblock Lora: Low-rank adaptation of large language models.
\newblock \emph{arXiv preprint arXiv:2106.09685}, 2021.

\bibitem[Hu et~al.(2023)Hu, Gao, Zhang, Sun, Zhang, and Bo]{hu2023animate}
Li Hu, Xin Gao, Peng Zhang, Ke Sun, Bang Zhang, and Liefeng Bo.
\newblock Animate anyone: Consistent and controllable image-to-video synthesis for character animation.
\newblock \emph{arXiv preprint arXiv:2311.17117}, 2023.

\bibitem[Huang et~al.(2024{\natexlab{a}})Huang, Wang, Zeng, Cao, Qi, Shi, Zha, and Zhang]{huang2024dreamwaltz}
Yukun Huang, Jianan Wang, Ailing Zeng, He Cao, Xianbiao Qi, Yukai Shi, Zheng-Jun Zha, and Lei Zhang.
\newblock Dreamwaltz: Make a scene with complex 3d animatable avatars.
\newblock \emph{Advances in Neural Information Processing Systems}, 36, 2024{\natexlab{a}}.

\bibitem[Huang et~al.(2024{\natexlab{b}})Huang, Yi, Xiu, Liao, Tang, Cai, and Thies]{huang2024tech}
Yangyi Huang, Hongwei Yi, Yuliang Xiu, Tingting Liao, Jiaxiang Tang, Deng Cai, and Justus Thies.
\newblock Tech: Text-guided reconstruction of lifelike clothed humans.
\newblock In \emph{2024 International Conference on 3D Vision (3DV)}, pages 1531--1542. IEEE, 2024{\natexlab{b}}.

\bibitem[Kerbl et~al.(2023)Kerbl, Kopanas, Leimk{\"u}hler, and Drettakis]{kerbl20233d}
Bernhard Kerbl, Georgios Kopanas, Thomas Leimk{\"u}hler, and George Drettakis.
\newblock 3d gaussian splatting for real-time radiance field rendering.
\newblock \emph{ACM Trans. Graph.}, 42\penalty0 (4):\penalty0 139--1, 2023.

\bibitem[Kim et~al.(2023)Kim, Gu, Park, Park, and Choo]{kim2023stableviton}
Jeongho Kim, Gyojung Gu, Minho Park, Sunghyun Park, and Jaegul Choo.
\newblock Stableviton: Learning semantic correspondence with latent diffusion model for virtual try-on.
\newblock \emph{arXiv preprint arXiv:2312.01725}, 2023.

\bibitem[Kingma(2013)]{kingma2013auto}
Diederik~P Kingma.
\newblock Auto-encoding variational bayes.
\newblock \emph{arXiv preprint arXiv:1312.6114}, 2013.

\bibitem[Lahner et~al.(2018)Lahner, Cremers, and Tung]{lahner2018deepwrinkles}
Zorah Lahner, Daniel Cremers, and Tony Tung.
\newblock Deepwrinkles: Accurate and realistic clothing modeling.
\newblock In \emph{Proceedings of the European conference on computer vision (ECCV)}, pages 667--684, 2018.

\bibitem[Lee et~al.(2022)Lee, Gu, Park, Choi, and Choo]{lee2022hrviton}
Sangyun Lee, Gyojung Gu, Sunghyun Park, Seunghwan Choi, and Jaegul Choo.
\newblock High-resolution virtual try-on with misalignment and occlusion-handled conditions.
\newblock In \emph{Proceedings of the European conference on computer vision (ECCV)}, 2022.

\bibitem[Loper et~al.(2023)Loper, Mahmood, Romero, Pons-Moll, and Black]{loper2023smpl}
Matthew Loper, Naureen Mahmood, Javier Romero, Gerard Pons-Moll, and Michael~J Black.
\newblock Smpl: A skinned multi-person linear model.
\newblock In \emph{Seminal Graphics Papers: Pushing the Boundaries, Volume 2}, pages 851--866. 2023.

\bibitem[Men et~al.(2020)Men, Mao, Jiang, Ma, and Lian]{men2020controllable}
Yifang Men, Yiming Mao, Yuning Jiang, Wei-Ying Ma, and Zhouhui Lian.
\newblock Controllable person image synthesis with attribute-decomposed gan.
\newblock In \emph{Proceedings of the IEEE/CVF conference on computer vision and pattern recognition}, pages 5084--5093, 2020.

\bibitem[Mir et~al.(2020)Mir, Alldieck, and Pons-Moll]{mir2020learning}
Aymen Mir, Thiemo Alldieck, and Gerard Pons-Moll.
\newblock Learning to transfer texture from clothing images to 3d humans.
\newblock In \emph{Proceedings of the IEEE/CVF Conference on Computer Vision and Pattern Recognition}, pages 7023--7034, 2020.

\bibitem[Morelli et~al.(2023)Morelli, Baldrati, Cartella, Cornia, Bertini, and Cucchiara]{morelli2023ladi}
Davide Morelli, Alberto Baldrati, Giuseppe Cartella, Marcella Cornia, Marco Bertini, and Rita Cucchiara.
\newblock Ladi-vton: Latent diffusion textual-inversion enhanced virtual try-on.
\newblock \emph{arXiv preprint arXiv:2305.13501}, 2023.

\bibitem[Oquab et~al.(2023)Oquab, Darcet, Moutakanni, Vo, Szafraniec, Khalidov, Fernandez, Haziza, Massa, El-Nouby, et~al.]{oquab2023dinov2}
Maxime Oquab, Timoth{\'e}e Darcet, Th{\'e}o Moutakanni, Huy Vo, Marc Szafraniec, Vasil Khalidov, Pierre Fernandez, Daniel Haziza, Francisco Massa, Alaaeldin El-Nouby, et~al.
\newblock Dinov2: Learning robust visual features without supervision.
\newblock \emph{arXiv preprint arXiv:2304.07193}, 2023.

\bibitem[Pavlakos et~al.(2019)Pavlakos, Choutas, Ghorbani, Bolkart, Osman, Tzionas, and Black]{pavlakos2019expressive}
Georgios Pavlakos, Vasileios Choutas, Nima Ghorbani, Timo Bolkart, Ahmed~AA Osman, Dimitrios Tzionas, and Michael~J Black.
\newblock Expressive body capture: 3d hands, face, and body from a single image.
\newblock In \emph{Proceedings of the IEEE/CVF conference on computer vision and pattern recognition}, pages 10975--10985, 2019.

\bibitem[Pons-Moll et~al.(2017)Pons-Moll, Pujades, Hu, and Black]{pons2017clothcap}
Gerard Pons-Moll, Sergi Pujades, Sonny Hu, and Michael~J Black.
\newblock Clothcap: Seamless 4d clothing capture and retargeting.
\newblock \emph{ACM Transactions on Graphics (ToG)}, 36\penalty0 (4):\penalty0 1--15, 2017.

\bibitem[Poole et~al.(2022)Poole, Jain, Barron, and Mildenhall]{poole2022dreamfusion}
Ben Poole, Ajay Jain, Jonathan~T Barron, and Ben Mildenhall.
\newblock Dreamfusion: Text-to-3d using 2d diffusion.
\newblock \emph{arXiv preprint arXiv:2209.14988}, 2022.

\bibitem[Ren et~al.(2022)Ren, Fan, Li, Liu, and Li]{ren2022neural}
Yurui Ren, Xiaoqing Fan, Ge Li, Shan Liu, and Thomas~H Li.
\newblock Neural texture extraction and distribution for controllable person image synthesis.
\newblock In \emph{Proceedings of the IEEE/CVF Conference on Computer Vision and Pattern Recognition}, pages 13535--13544, 2022.

\bibitem[Rombach et~al.(2022)Rombach, Blattmann, Lorenz, Esser, and Ommer]{rombach2022high}
Robin Rombach, Andreas Blattmann, Dominik Lorenz, Patrick Esser, and Bj{\"o}rn Ommer.
\newblock High-resolution image synthesis with latent diffusion models.
\newblock In \emph{Proceedings of the IEEE/CVF Conference on Computer Vision and Pattern Recognition}, pages 10684--10695, 2022.

\bibitem[Ruiz et~al.(2022)Ruiz, Li, Jampani, Pritch, Rubinstein, and Aberman]{ruiz2022dreambooth}
Nataniel Ruiz, Yuanzhen Li, Varun Jampani, Yael Pritch, Michael Rubinstein, and Kfir Aberman.
\newblock Dreambooth: Fine tuning text-to-image diffusion models for subject-driven generation.
\newblock \emph{arXiv preprint arXiv:2208.12242}, 2022.

\bibitem[Santesteban et~al.(2021)Santesteban, Thuerey, Otaduy, and Casas]{santesteban2021self}
Igor Santesteban, Nils Thuerey, Miguel~A Otaduy, and Dan Casas.
\newblock Self-supervised collision handling via generative 3d garment models for virtual try-on.
\newblock In \emph{Proceedings of the IEEE/CVF Conference on Computer Vision and Pattern Recognition}, pages 11763--11773, 2021.

\bibitem[Santesteban et~al.(2022)Santesteban, Otaduy, Thuerey, and Casas]{santesteban2022ulnef}
Igor Santesteban, Miguel Otaduy, Nils Thuerey, and Dan Casas.
\newblock Ulnef: Untangled layered neural fields for mix-and-match virtual try-on.
\newblock \emph{Advances in Neural Information Processing Systems}, 35:\penalty0 12110--12125, 2022.

\bibitem[Sohl-Dickstein et~al.(2015)Sohl-Dickstein, Weiss, Maheswaranathan, and Ganguli]{sohl2015deep}
Jascha Sohl-Dickstein, Eric Weiss, Niru Maheswaranathan, and Surya Ganguli.
\newblock Deep unsupervised learning using nonequilibrium thermodynamics.
\newblock In \emph{International Conference on Machine Learning}, pages 2256--2265. PMLR, 2015.

\bibitem[Song et~al.(2020)Song, Meng, and Ermon]{song2020denoising}
Jiaming Song, Chenlin Meng, and Stefano Ermon.
\newblock Denoising diffusion implicit models.
\newblock \emph{arXiv preprint arXiv:2010.02502}, 2020.

\bibitem[Song and Ermon(2019)]{song2019generative}
Yang Song and Stefano Ermon.
\newblock Generative modeling by estimating gradients of the data distribution.
\newblock \emph{Advances in Neural Information Processing Systems}, 32, 2019.

\bibitem[Wang et~al.(2024{\natexlab{a}})Wang, Zhang, Di, Zhang, and Zuo]{wang2024mv}
Haoyu Wang, Zhilu Zhang, Donglin Di, Shiliang Zhang, and Wangmeng Zuo.
\newblock Mv-vton: Multi-view virtual try-on with diffusion models.
\newblock \emph{arXiv preprint arXiv:2404.17364}, 2024{\natexlab{a}}.

\bibitem[Wang et~al.(2024{\natexlab{b}})Wang, Fang, Zhang, Xie, and Tian]{wang2024gaussianeditor}
Junjie Wang, Jiemin Fang, Xiaopeng Zhang, Lingxi Xie, and Qi Tian.
\newblock Gaussianeditor: Editing 3d gaussians delicately with text instructions.
\newblock In \emph{Proceedings of the IEEE/CVF Conference on Computer Vision and Pattern Recognition}, pages 20902--20911, 2024{\natexlab{b}}.

\bibitem[Wu et~al.(2024)Wu, Bian, Li, Wang, Reid, Torr, and Prisacariu]{wu2024gaussctrl}
Jing Wu, Jia-Wang Bian, Xinghui Li, Guangrun Wang, Ian Reid, Philip Torr, and Victor~Adrian Prisacariu.
\newblock Gaussctrl: multi-view consistent text-driven 3d gaussian splatting editing.
\newblock \emph{arXiv preprint arXiv:2403.08733}, 2024.

\bibitem[Wu et~al.(2023)Wu, Ge, Wang, Lei, Gu, Shi, Hsu, Shan, Qie, and Shou]{wu2023tune}
Jay~Zhangjie Wu, Yixiao Ge, Xintao Wang, Stan~Weixian Lei, Yuchao Gu, Yufei Shi, Wynne Hsu, Ying Shan, Xiaohu Qie, and Mike~Zheng Shou.
\newblock Tune-a-video: One-shot tuning of image diffusion models for text-to-video generation.
\newblock In \emph{Proceedings of the IEEE/CVF International Conference on Computer Vision}, pages 7623--7633, 2023.

\bibitem[Xie et~al.(2024)Xie, Dong, Gao, Ma, and Liang]{xie2024dreamvton}
Zhenyu Xie, Haoye Dong, Yufei Gao, Zehua Ma, and Xiaodan Liang.
\newblock Dreamvton: Customizing 3d virtual try-on with personalized diffusion models.
\newblock \emph{arXiv preprint arXiv:2407.16511}, 2024.

\bibitem[Xiong et~al.(2024)Xiong, Li, Liu, Liao, Hu, Zhu, Ning, Qiu, Wang, Wang, et~al.]{xiong2024mvhumannet}
Zhangyang Xiong, Chenghong Li, Kenkun Liu, Hongjie Liao, Jianqiao Hu, Junyi Zhu, Shuliang Ning, Lingteng Qiu, Chongjie Wang, Shijie Wang, et~al.
\newblock Mvhumannet: A large-scale dataset of multi-view daily dressing human captures.
\newblock In \emph{Proceedings of the IEEE/CVF Conference on Computer Vision and Pattern Recognition}, pages 19801--19811, 2024.

\bibitem[Xu et~al.(2024{\natexlab{a}})Xu, Nie, Liu, Liu, Kautz, Wang, and Vahdat]{xu2024camco}
Dejia Xu, Weili Nie, Chao Liu, Sifei Liu, Jan Kautz, Zhangyang Wang, and Arash Vahdat.
\newblock Camco: Camera-controllable 3d-consistent image-to-video generation.
\newblock \emph{arXiv preprint arXiv:2406.02509}, 2024{\natexlab{a}}.

\bibitem[Xu et~al.(2024{\natexlab{b}})Xu, Gu, Chen, and Chen]{xu2024ootdiffusion}
Yuhao Xu, Tao Gu, Weifeng Chen, and Chengcai Chen.
\newblock Ootdiffusion: Outfitting fusion based latent diffusion for controllable virtual try-on.
\newblock \emph{arXiv preprint arXiv:2403.01779}, 2024{\natexlab{b}}.

\bibitem[Yang et~al.(2022)Yang, Yu, and Liu]{Yang_2022_CVPR}
Han Yang, Xinrui Yu, and Ziwei Liu.
\newblock Full-range virtual try-on with recurrent tri-level transform.
\newblock In \emph{Proceedings of the IEEE/CVF Conference on Computer Vision and Pattern Recognition (CVPR)}, pages 3460--3469, 2022.

\bibitem[Yu et~al.(2021)Yu, Zheng, Guo, Liu, Dai, and Liu]{yu2021function4d}
Tao Yu, Zerong Zheng, Kaiwen Guo, Pengpeng Liu, Qionghai Dai, and Yebin Liu.
\newblock Function4d: Real-time human volumetric capture from very sparse consumer rgbd sensors.
\newblock In \emph{Proceedings of the IEEE/CVF conference on computer vision and pattern recognition}, pages 5746--5756, 2021.

\bibitem[Zhang et~al.(2021)Zhang, Li, Lai, and Yang]{zhang2021pise}
Jinsong Zhang, Kun Li, Yu-Kun Lai, and Jingyu Yang.
\newblock Pise: Person image synthesis and editing with decoupled gan.
\newblock In \emph{Proceedings of the IEEE/CVF Conference on Computer Vision and Pattern Recognition}, pages 7982--7990, 2021.

\bibitem[Zhang and Agrawala(2023)]{zhang2023adding}
Lvmin Zhang and Maneesh Agrawala.
\newblock Adding conditional control to text-to-image diffusion models.
\newblock \emph{arXiv preprint arXiv:2302.05543}, 2023.

\bibitem[Zhang et~al.(2023)Zhang, Wei, Jiang, Zhang, Zuo, and Tian]{zhang2023controlvideo}
Yabo Zhang, Yuxiang Wei, Dongsheng Jiang, Xiaopeng Zhang, Wangmeng Zuo, and Qi Tian.
\newblock Controlvideo: Training-free controllable text-to-video generation.
\newblock \emph{arXiv preprint arXiv:2305.13077}, 2023.

\bibitem[Zhao et~al.(2021)Zhao, Xie, Kampffmeyer, Dong, Han, Zheng, Zhang, and Liang]{zhao2021m3d}
Fuwei Zhao, Zhenyu Xie, Michael Kampffmeyer, Haoye Dong, Songfang Han, Tianxiang Zheng, Tao Zhang, and Xiaodan Liang.
\newblock M3d-vton: A monocular-to-3d virtual try-on network.
\newblock In \emph{Proceedings of the IEEE/CVF International Conference on Computer Vision}, pages 13239--13249, 2021.

\bibitem[Zhu et~al.(2023)Zhu, Yang, Zhu, Reda, Chan, Saharia, Norouzi, and Kemelmacher-Shlizerman]{zhu2023tryondiffusion}
Luyang Zhu, Dawei Yang, Tyler Zhu, Fitsum Reda, William Chan, Chitwan Saharia, Mohammad Norouzi, and Ira Kemelmacher-Shlizerman.
\newblock Tryondiffusion: A tale of two unets.
\newblock In \emph{Proceedings of the IEEE/CVF Conference on Computer Vision and Pattern Recognition}, pages 4606--4615, 2023.

\bibitem[Zhuang et~al.(2023)Zhuang, Wang, Lin, Liu, and Li]{zhuang2023dreameditor}
Jingyu Zhuang, Chen Wang, Liang Lin, Lingjie Liu, and Guanbin Li.
\newblock Dreameditor: Text-driven 3d scene editing with neural fields.
\newblock In \emph{SIGGRAPH Asia 2023 Conference Papers}, pages 1--10, 2023.

\bibitem[Zhuang et~al.(2024{\natexlab{a}})Zhuang, Kang, Bao, Lin, and Li]{zhuang2024dagsm}
Jingyu Zhuang, Di Kang, Linchao Bao, Liang Lin, and Guanbin Li.
\newblock Dagsm: Disentangled avatar generation with gs-enhanced mesh.
\newblock \emph{arXiv preprint arXiv:2411.15205}, 2024{\natexlab{a}}.

\bibitem[Zhuang et~al.(2024{\natexlab{b}})Zhuang, Kang, Cao, Li, Lin, and Shan]{zhuang2024tip}
Jingyu Zhuang, Di Kang, Yan-Pei Cao, Guanbin Li, Liang Lin, and Ying Shan.
\newblock Tip-editor: An accurate 3d editor following both text-prompts and image-prompts.
\newblock \emph{arXiv preprint arXiv:2401.14828}, 2024{\natexlab{b}}.

\end{thebibliography}
}

\clearpage
\setcounter{page}{1}
\maketitlesupplementary
\appendix

\noindent Appendix~\ref{sec:3DGS} introduces the preliminaries of 3DGS. The detailed formulations of the two quantitative metrics are presented in Appendix~\ref{sec:metrics}. Additionally, Appendix~\ref{sec:postprocess} outlines the post-processing techniques applied to ensure the preservation of human characteristics in image editing. Appendix~\ref{sec:limitation} elaborates on the failure cases and proposes a mitigation strategy to address it. Finally, Appendix~\ref{sec:more_cases} showcases additional VTON results, including those from a real 3D scene used in GaussianVTON~\cite{chen2024gaussianvton}.


\section{3D Representation: Gaussian Splatting}
\label{sec:3DGS}
3D Gaussian Splatting (3DGS)~\cite{kerbl20233d} has emerged as a prominent technique in 3D reconstruction due to its ability to render high-quality scenes in real-time. Unlike traditional point cloud based methods, which directly represent scenes as discrete points, 3DGS models each point as a continuous Gaussian function $g_i$:
\begin{equation}
g_i(
    \mathbf{\mathit{x}}; 
    \mathbf{\mathit{\mu}}_i, 
    \mathbf{\mathit{\Sigma}}_i
)
= e^{
    - \frac{1}{2}
    (\mathbf{\mathit{x}} - \mathbf{\mathit{\mu}}_i)^\top
    \mathbf{\mathit{\Sigma}_i}
    (\mathbf{\mathit{x}} - \mathbf{\mathit{\mu}}_i) 
}
,
\end{equation}
where $\mathbf{\mathit{x}}$ is the position vector of $g_i$,
$\mathbf{\mathit{\mu}}_i \in \mathbb{R}^3$ and $\mathbf{\mathit{\Sigma}}_i \in \mathbb{R}^{3 \times 3}$ are $g_i$'s mean and covariance matrix, respectively.
Then, $g_i$ is projected onto a 2D image plane to facilitate rendering.
This projection yields a new mean vector $\mathbf{\mathit{\mu_i}}^\prime \in \mathbb{R}^2$ and an updated covariance matrix $\mathbf{\mathit{\Sigma}}^{\prime}_i \in \mathbb{R}^{2 \times 2}$ defined as:
\begin{equation}
\mathbf{\mathit{\mu_i}}^\prime
= \mathbf{\mathit{K}} \mathbf{\mathit{T}} [\mathbf{\mathit{\mu_i}}^\top, 1]^\top
,
\mathbf{\mathit{\Sigma}}^{\prime}_i
= \mathbf{\mathit{J}}  \mathbf{\mathit{T}} \mathbf{\mathit{\Sigma}}_i \mathbf{\mathit{T}}^\top \mathbf{\mathit{J}}^\top,
\end{equation}
where $\mathbf{\mathit{J}}$ is the Jacobian matrix derived from the affine approximation of the perspective projection, $\mathbf{\mathit{T}}$ and $\mathbf{\mathit{K}}$ denote the extrinsic and intrinsic matrices, respectively. Given the color $c_i$ and opacity $\alpha_i$ at the Gaussian center point, the rendered color at a 2D pixel $\mathbf{\mathit{p}}$ is calculated as follows:
\begin{equation}
\begin{aligned}
\mathbf{\mathit{C}}_{\mathbf{\mathit{p}}}
&= \sum_{i=1}^{N}{
    \mathit{\alpha}_i
    \mathit{c}_i
    \mathit{T}_i
    \mathit{g}_i(
        \mathbf{\mathit{p}};
        \mathbf{\mathit{\mu}}^{\prime}_i,
        \mathbf{\mathit{\Sigma}}^{\prime}_i
    )
}
\\
\mathit{T}_i
&= \prod_{j=1}^{i-1}{
    (
        1 - \mathit{\alpha}_j
        \mathit{g}_j(
            \mathbf{\mathit{p}};
            \mathbf{\mathit{\mu}}^{\prime}_j,
            \mathbf{\mathit{\Sigma}}^{\prime}_j
        )
    )
}
,
\end{aligned}
\end{equation}
where $\mathit{T}_i$ denotes the cumulative transmission along the ray.

\section{Metrics}
\label{sec:metrics}
In the quantitative evaluation, we employ two metrics: 
\begin{itemize}
    \item Average DINO Similarity~\cite{zhuang2024tip}, which measures the alignment between the garment image and the edited 3D human.
    \item CLIP Directional Consistency Score~\cite{haque2023instruct}, which evaluates multi-view consistency.
\end{itemize}
Specifically, given an edited 3D human (after VTON), 120 views are uniformly projected around its central axis. These views are divided into three categories based on orientation: $S_f$, $S_b$, and $S_s$, corresponding to 40 front views, 40 back views, and 40 side views, respectively.  
Let $D(\cdot)$ represent the normalized DINO embedding and $C(\cdot)$ denote the normalized CLIP embedding. Using these, we formally define the two metrics as follows:  

\begin{equation}
{\small
\begin{aligned}
    \text{DINO}_{sim}&=\frac{1}{80}(\sum_{i\in S_{f}}D(g_f)\cdot D(e_i)+\sum_{i\in S_{b}}D(g_b)\cdot D(e_i))\\
    \text{CLIP}_{cons}&=\frac{1}{120}\sum_i (C(e_i)-C(o_i))\cdot (C(e_{i+1})-C(o_{i+1}))
    \label{eq:metrics}
\end{aligned}}
\end{equation}
\noindent
where $e_i$, $e_{i+1}$ and $o_i$, $o_{i+1}$ denotes the two consecutive novel views from the edited 3DGS and the original 3DGS, respectively.

\section{Post-processing}
\label{sec:postprocess}
The clothing-agnostic maps $\mathbf{A}$ often mask parts of the face and hair, particularly for females. Due to the inherent properties of the diffusion model, it is unable to fully restore the intricate details of these masked regions. To ensure high-fidelity preservation of human characteristics, we apply a post-processing step where, after editing the rendered views, we ``copy'' the face and hair from the original image $o$ onto the edited image $e$. Specifically, let $m$ represent the region corresponding to the face and hair, which can be extracted from the parsed map during pre-processing, we implement post-processing as: 
\begin{equation}
    e = (1-m)\cdot e + m\cdot o
\end{equation}

\begin{figure}[h]
\centering
\includegraphics[width=1.0\linewidth]{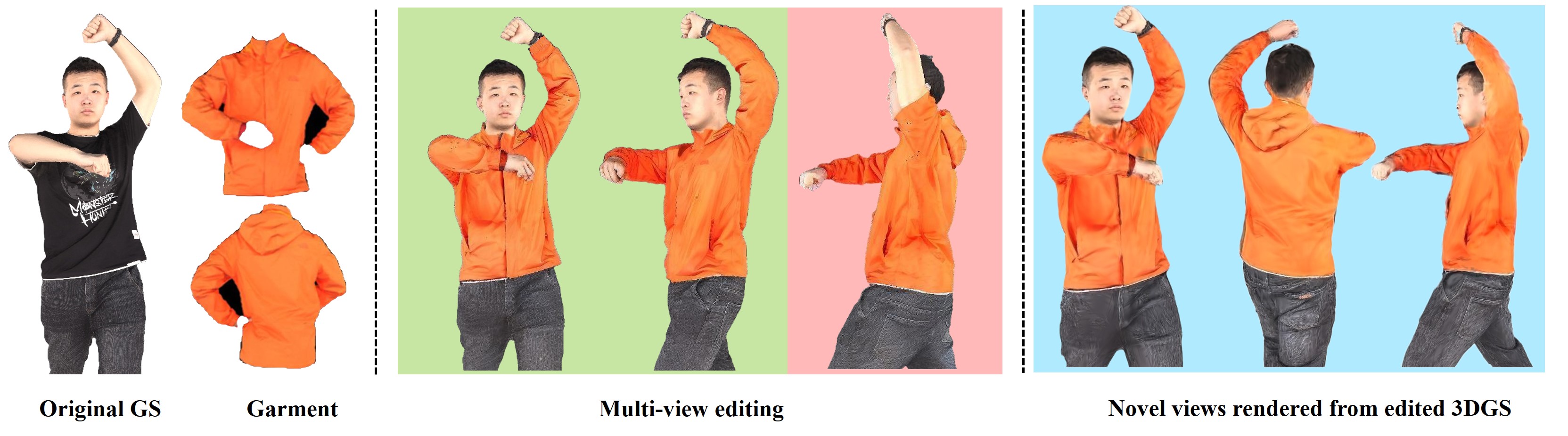}
\vspace{-12pt}
\caption{Our multi-view editing may fail in certain views with complex poses (\textcolor{red}{red} box in pink background) but these views can be automatically discarded to mitigate their impact on 3D VTON (\textcolor{blue}{blue} background).}
\label{fig:limitations}
\vspace{-12pt}
\end{figure}

\section{Limitations}
\label{sec:limitation}
As shown in Fig.~\ref{fig:limitations}, our method may fail in certain views with complex postures. To address this, we use Z-Score Normalization to automatically identify and discard problematic views based on the view reconstruction loss during the process of lifting multiple views to 3D space, mitigating their adverse impact. 

\section{Additional Visualization Results}
\label{sec:more_cases}
\begin{figure*} [ht]
	\begin{center}
		\includegraphics[width=1.0\linewidth]{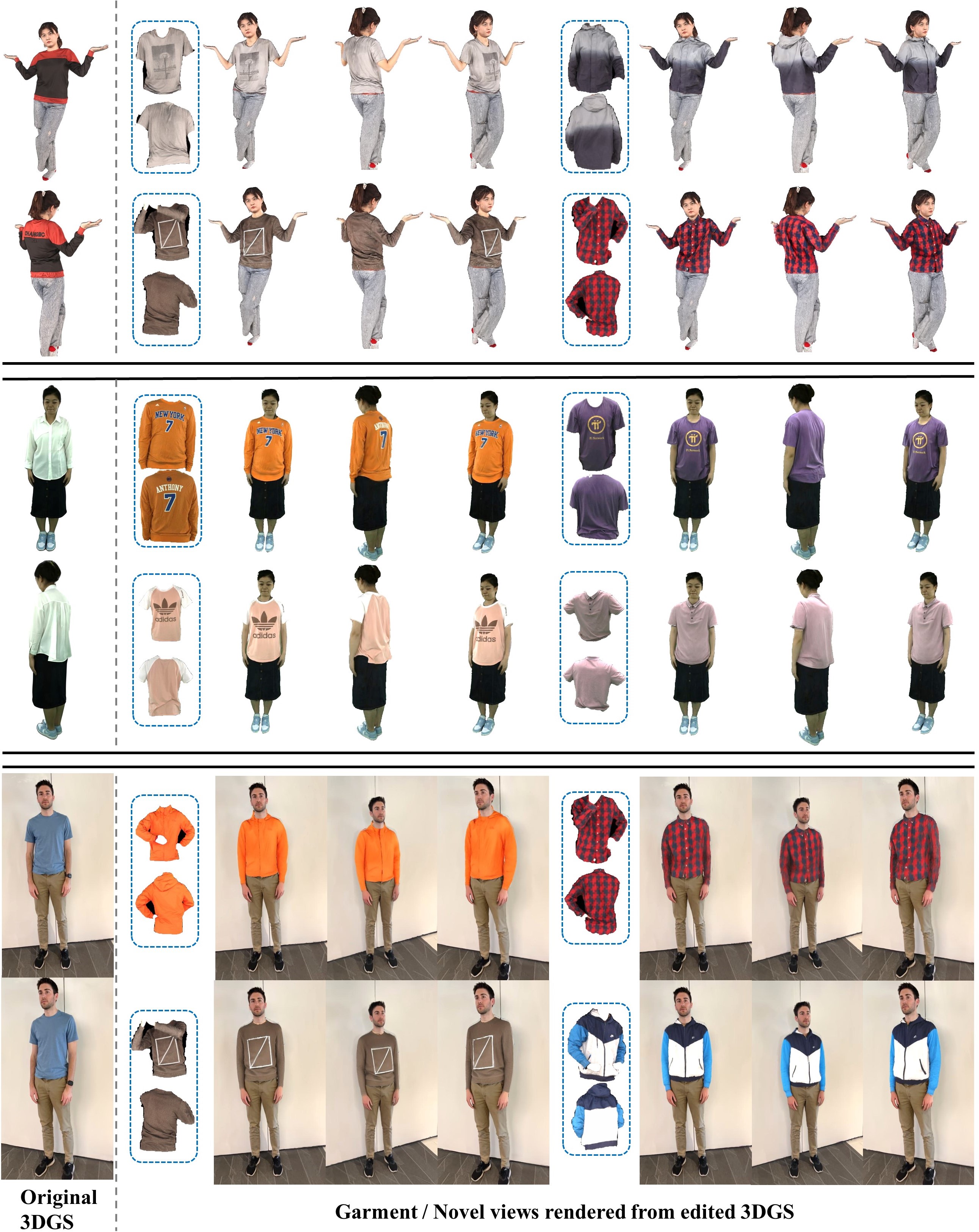} 
	\end{center}
        \vspace{-10pt}
    \caption{\textbf{Additional visualization results.} The first, middle, and last two rows show results on Thuman2.0, MVHumanNet, and a real 3D scene used in GaussianVTON, respectively.
    }
 \label{fig:add_cases}
 \vspace{-11pt}
\end{figure*}

Fig.~\ref{fig:add_cases} illustrates additional VTON results. 
The first two rows showcase results from the THuman2.0 dataset; the middle two rows showcase results from the MVHumanNet dataset. To further demonstrate the effectiveness of our method, we apply it on a real 3D scene used in GaussianVTON~\cite{chen2024gaussianvton}. The last two rows in Fig.~\ref{fig:add_cases} illustrate these VTON results with the model trained on Thuman2.0 dataset. Despite the data gap, including w/wo background and unseen camera poses, our method exhibits robust performance and preserves the details of the clothing well.


\end{document}